%% file: main.tex
\documentclass[letterpaper, 10pt, conference]{ieeeconf}
\IEEEoverridecommandlockouts
\usepackage{amsmath,amssymb,amsfonts}
\usepackage{textcomp}
\usepackage{xcolor}
\usepackage{booktabs}
\usepackage{algorithm}
\usepackage{algpseudocode}
\usepackage{xspace}
\usepackage{cleveref}
\usepackage{graphicx}
\usepackage{wrapfig}
\usepackage[normalem]{ulem}
\usepackage{booktabs, multirow}
\usepackage{url}
\usepackage{subcaption}

\begin{document}

\input{notations}

\title{RE4: Transformation-aware Imitation of\\Object Interactions Using Manipulation Modes%
\thanks{This work has been submitted to the IEEE for possible publication.
Copyright may be transferred without notice, after which this version may
no longer be accessible.}}

\author{
\authorblockN{Arsh Chawla}
\authorblockA{Australian National University}
\and
\authorblockN{Rahul Shome}
\authorblockA{Australian National University}
}

\maketitle

\begin{abstract}
    Object interaction tasks have been a focus of advances in imitation learning. 
    End-to-end methods, 
    dominated by 
    generative variants 
    have shown leaps in performance but tend to be black boxes. 
    Recent indications hint at demonstration lookup as the operating process underneath generative policy performance.
    In this paper, we revisit a few modern imitation learning benchmarks for object interactions, with the aim of composing a simple framework from first principles. We repurpose theories of manipulation modes and apply them to retrieve and transform demonstration datasets, creating a minimal strategy that preserves performance and robustness while remaining interpretable. 
    We identify pose estimation as a key module that governs information about the object.
    This informs a manipulation mode-aware retrieval of a demonstration, a mode-aware transformation, a replan step that connects to the retrieval point while preserving mode constraints, returning a transformed demonstration for rollout. These compose four key steps of the proposed \acronym framework, evaluated over state-based and image-based benchmarks in simulated Push-T, Robomimic, and MimicGen as well as a real robot experiment. Robustness is measured in an adversarial study that evaluates sparse data regimes, a study with visual distractors, and the addition of an obstructing collision geometry.   
    \acronym outperforms baselines, while notably preserving interpretability.
    The results overwhelmingly motivate the benefits of identifying simple interpretable building blocks to learn manipulation skills for object interactions.          
\end{abstract}


\section{Introduction}
\label{sec:introduction}

Manipulation~\cite{mason2018toward} is a well-studied area that has historically been a key driving application in robotics. 
Learning behaviour from expert demonstrations has shown promise for complex, contact-rich manipulation \cite{correia2023surveydemonstrationlearning}. Recent generative policies, diffusion and flow-matching models, now set the state of the art by capturing the multimodal, discontinuous structure of expert behaviour that classical explicit policies are not expressive enough to represent \cite{chi2024diffusionpolicy, wang2024equivariantdiffusionpolicy}. That expressiveness, however, 
comes at the expense of
compute, demonstration budget, and interpretability, with
the resulting policy effectively becoming a black box \cite{pmlr-v305-jiang25a, wang2024equivariantdiffusionpolicy}. VLAs represent an even stronger shift in this trend~\cite{pmlr-v270-kim25c}. The question arises whether we can find a balance with interpretable yet performant composition of building blocks in these tasks.

The memorization characteristics of generative models for computer vision tasks have been investigated for generalization and robustness~\cite{bonnaire2026diffusion}. Recent analyses suggest much of this machinery may be unnecessary for robotic action trajectories, arguing that diffusion policies largely \emph{memorize} their demonstrations and replay the nearest one at inference~\cite{2025arXiv250505787H}---reframing retrieval, rather than generation, as the operative mechanism. Yet existing retrieval- and object-centric methods each recover only a fragments of what may be needed---nearest-neighbour reuse, or a single per-episode pose transfer~\cite{vitiello2023oneshotimitationlearningpose, ICLR2026_bf235a1d}. We ask -- \textit{if retrieval explains performance of black box methods, in part, can we design a retrieval-based method to achieve similar performance? What other design choice explains an effective policy framework?}

\begin{figure}[t]
    \centering
    \includegraphics[width=1.0\linewidth]{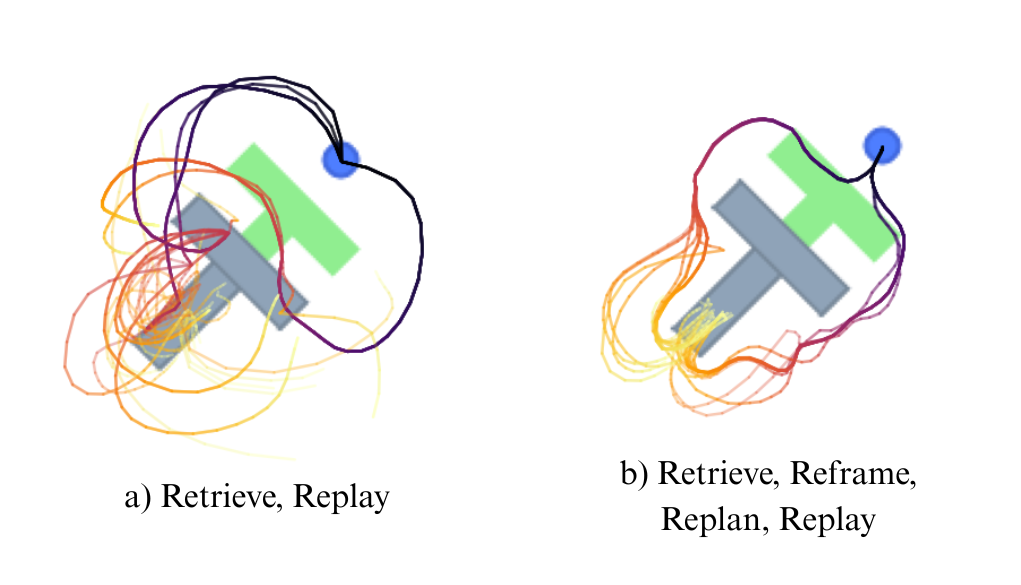}
    \caption{
    \textbf{\Expansion:} 
    On the left, the ablation using only the \textit{retrieval} and \textit{replay} 
    rollouts fail to make progress in the task.
    All four steps are used in the rollouts visualized on the right.
    The right image visualizes multiple $\epsilon$-near candidates for retrieval in \acronym.
    Notably, the expected accurately transformed multimodality emerges.
    }
    \label{fig:placeholder}
\end{figure}

We identify that behaviour switching at the contact points,
where the relationship between gripper and object changes discontinuously,
forms critical points in tasks. This hints at the object pose being a key element in the problem. Our observations correspond to the well-studied theories of manipulation modes~\cite{kuffner2016motion,hauser2010multi}. Modeling contact modalities of manipulation closely relates to identifying the topologies of the task constraints~\cite{kingston2019decoupling}, differing in analysis when non-prehensile manipulation is considered~\cite{dogar2011framework,vieira2022persistent}. The connectivity of these spaces is also understood to affect how information can be shared between different modes and contacts~\cite{kingston2021using}. Motion controllers~\cite{ijspeert2013dynamical} learned from demonstrations have also informed task constraints in planning~\cite{sobti2021sampling}. Here, we identify our key question -- \emph{can theories of multimodal manipulation inform the pieces critical to building a simple framework for imitating object interactions from demonstration?} We stress on simplicity because we aim for a minimal framework sufficient for explaining and achieving performance in object interaction imitation learning settings.
We focus on task-specific imitation learning problems, where demonstrations are used to learn task-specific policies.

\begin{figure*}[t]
    \centering
    \includegraphics[width=0.95\linewidth]{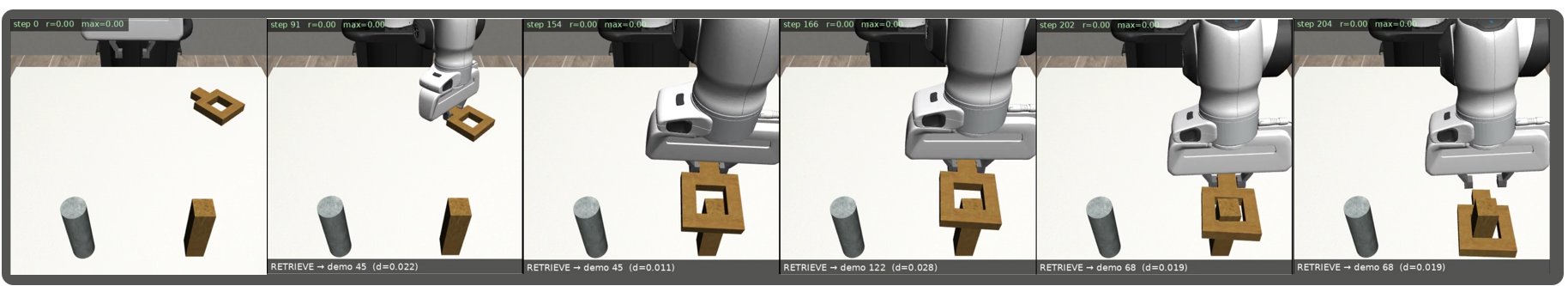}
    \caption{An interpretable sequence of \acronym rollouts in a Robomimic Square benchmark, indicating the demonstrations (45, 122, 68)  that were retrieved (then reframed, replanned, and replayed).}
    \label{fig:square}
\end{figure*}

In this paper we compose a handful of simple, lightweight or pretrained~\cite{Wen_2024_CVPR,any6d}, and interpretable modules (Fig~\ref{fig:placeholder})---transformation-aware retrieval, object-relative reframing, planned bridging, and replay, each conditioned on the underlying \emph{manipulation mode} informed by an object pose estimate. 
Building on principles drawn from manipulation theory, we propose a method that matches or exceeds baselines at performance and generalization
while remaining exactly interpretable. 
We identify a key component as the pose estimator. 
We evaluate task-specific lightweight trained estimators, as well as foundational pre-trained ones~\cite{Wen_2024_CVPR,any6d}.
Using the composable framework, each component while simple, together prove sufficient to achieve robust, interpretable performance
(see Fig \ref{fig:square} for exact interpretability).  

Our contributions are: 
(i) identifying key components of a framework \acronym for object interaction imitation as retrieval, reframing, replanning, replay, and pose estimation; 
(ii) demonstrating
performance benefits of \acronym 
compared to baselines in state and image-based tasks in simulation and on the real robot
(iii) demonstrating generalization and robustness of \acronym in sparse data regimes, with visual distractors, and for downstream filters like  collision checking. 

The results are a promising motivation to revisit the core and simple building blocks for learning manipulation skills. Inspecting the problem through a principled theoretical lens allows us to composing interpretable policies that do not need to trade off performance or robustness for interpretability. 




\section{Related Work}
\label{sec:related}

\textbf{Generative policies.} Inspired by image generation, diffusion and flow-matching generative models have shown strong performance in behavioural cloning across contact-rich manipulation tasks \cite{pmlr-v305-jiang25a, chi2024diffusionpolicy, 2025arXiv250505787H}---requiring only expert demonstration data. A downside to generative policies are the associated computational costs, and lack of interpretablity \cite{shen2025efficientdiffusionmodelssurvey}. Efficiency-focused variants reduce the cost---distilling to few-step generation \cite{prasad2024consistencypolicyacceleratedvisuomotor}, enforcing $SO(2)$ symmetry in the denoiser \cite{wang2024equivariantdiffusionpolicy}, or flowing
in action rather than action-sequence space \cite{pmlr-v305-jiang25a}---but do not solve issues entirely. VLAs~\cite{pmlr-v270-kim25c} have pushed this paradigm to cross-task foundational regimes.

\textbf{Retrieval-based imitation learning.} He et. al~\cite{2025arXiv250505787H} argue diffusion policies largely \emph{memorise} their demonstrations, and propose a latent space based Action Lookup Table that operates at a fraction of the cost. Retrieval based methods predate generative policies, VINN
\cite{Pari-RSS-22} predicts actions as a weighted average over the nearest demonstration frames in a learned embedding, but replays them \emph{as recorded} and is thus bound to
demonstration coverage. Palo et. al~\cite{2024IRAL}
decompose reuse into retrieval, alignment, and replay---the closest structure to
RE4---but their alignment is a \emph{learned} visual servo, with no explicit object-frame transform and no notion of manipulation mode.

\textbf{Object-centric imitation learning.}
A long line of structured IL encodes demonstrations as reusable, object-relative motion models---task-parameterized skills in object frames \cite{huang2018generalizedtaskparameterizedskilllearning}, and keypoint groundings \cite{franzese2025generalizablemotionpolicieskeypoint}---treating object pose as the anchor that \emph{transports} a demonstration. One-shot imitation \cite{vitiello2023oneshotimitationlearningpose} transports a demonstrated end-effector trajectory by an estimated relative object pose, with no notion of \textit{manipulation mode}, while GPI \cite{ICLR2026_bf235a1d} composes
flows over the nearest demonstrations rather than applying an explicit transform. The ingredients of RE4---nearest-neighbour retrieval, object-pose transfer, and pose-aware reuse---appear separately.

\textbf{Pose Estimation.} Object pose is a powerful conditioning signal for manipulation across retrieval and task-parameterised skill learning \cite{vitiello2023oneshotimitationlearningpose, huang2018generalizedtaskparameterizedskilllearning, ICLR2026_bf235a1d}, but training pose estimators without ground-truth supervision is non-trivial. General-purpose 6-DoF pose estimators exist \cite{Wen_2024_CVPR} but often require CAD models and object segmentation. Recovering the manipulated object's pose \emph{directly and only from the demonstration data} is largely unexplored, even though self-supervised pose estimation is well established in adjacent domains such as human-body pose \cite{sosa2023selfsupervised3dhumanpose}. RE4 proposes lightweight, self-supervised estimator trained on demonstrations alone (\cref{sec:pose-estimation}), while also ablating performance under foundational models.

\textbf{Manipulation.} A widely explored problem area, manipulation~\cite{mason2018toward} has presented principled modelling of the underlying search spaces as modes~\cite{kuffner2016motion,hauser2010multi}. Motion planning with constraints~\cite{kingston2019decoupling}, distinguishes transit and transfer, as well as contact parameters. Sequential manipulation expresses task and motion topologies~\cite{pan2024task} where the transition of modes inform careful consideration of neighbourhoods~\cite{kingston2021using}. These contact topologies and constraints change when non-prehensile interactions are introduced~\cite{dogar2011framework, vieira2022persistent, shome2019towards}. Imitation learning takes a data-centric approach to modelling manipulation constraints. The current work remains data-driven, while composing interpretable modules based on object pose, transformations, and manipulation modes.

\section{Preliminaries}
In this section, we present some preliminaries for contact-rich manipulation imitation. 

\paragraph{Object Manipulation} A robotic manipulator has an end-effector (eef)link $\eef$, describing an end-effector pose $\robotstate$ in a task space $\taskspace$. An end-effector configuration can describe its $SE(3)$ pose, i.e., $\robotstate \in \taskspace \subset SE(3)$ (for planar problems, the corresponding spaces may typically be $SE(2)$). The workspace of the robot is occupied by a target object at its pose $\object \in SE(3)$. The workspace is also occupied by other geometries. A sequence of end-effector configurations can describe a continuous motion for the manipulator, $\traj: [0,1] \rightarrow \taskspace$ where such a motion can go from a start configuration $\robotstate_{\mathrm{start}} = \traj(0)$ to some end configuration $\robotstate_{\mathrm{end}} = \traj(1)$. Next we carefully define manipulation modes necessary for the modeling sequential interactions with an object.

\textit{Manipulation Modes:} The combination of the end-effector and actuators, along with the object's properties express manipulation affordances. These correspond to task-relevant object interactions or actions. For the purposes of affordances (like grasping, pushing) defining a contact is necessary in terms of the object-centric relative transformation from the object to the eef $\transform{\object}{\robotstate}$. A wide category of these contact-rich interactions readily fall under the two manipulation modes - transit and transfer. Transit is defined as end-effector and object configurations from where the relative pose of the end-effector and object may change, i.e., the robot can move freely without the object in static, rigid contact. Transfer corresponds to configurations where the relative pose remains fixed, i.e., the object moves with the robot. The mode is then one of, $\mode = \{\transit,\,\transfer\} \in \mathcal{M}$. Consider an action changing the eef and object, $(\robotstate_i, \object_t) \overset{\action_t}{\rightarrow} (\robotstate_{t+1}, \object_{t+1})$. In the $\transit$ mode, $\| \robotstate_i, \robotstate_{t+1} \| - \| \object_i, \object_{t+1} \| \geq \epsilon$. In the $\transfer$, $\| \robotstate_i, \robotstate_{t+1} \| - \| \object_i, \object_{t+1} \| < \epsilon$ for some small $\epsilon\geq0$. The modes may switch at mode transitions $\transition$ that correspond to points of affordance contact like at grasps or placements. This means that $ \transform{\object^\times}{\robotstate^\times}$, which defines the relative transform at the start of a $\transfer$, remains preserved during transfer (in prehensile tasks).

\paragraph{Imitation Learning} A demonstration is a sequence of $N$ eef poses, observations, and actions 
$\demo = \{ \langle \robotstate_i, \obs_i, \action_i \rangle \}_{i=1}^H$. A set of $N$ demonstrations are provided for a task, $\demoset = \{ \demo_1, \cdots \demo_N \}$. Imitation learning computes a control policy that proposes an action given the eef pose and the observation.
$\action_t = \Policy_{\demoset}(\robotstate_t, \obs_t).$

\subsection{Foundations}
\label{sec:foundations}
Now we describe some necessary concepts that will be used to devise the framework.

\textit{Demonstration as Task Constraint: } A demonstration $\demo$ of $N$ steps, can be associated with a sequence of manipulation modes, $\mode_1\cdots \mode_N$. This may naturally form segments of consecutive modes of the same type, till it switches to a different mode. Each demonstration, $\demo$ corresponds to a sequence eef poses representing a discretization of the continuous end-effector trajectory for the task. Define a mode-augmented trajectory $\mtraj = [ ( \robotstate_1, \object_1, \mode_1 ), \cdots  ( \robotstate_N, \object_N, \mode_N )]$. Given a manipulation task, the available model for the task objective and constraints is $\demoset$, describing for a $\robotstate$ and $\obs$, a mode-augmented trajectory $\mtraj$ that \textit{solves} the task. 

For matching consecutive modes, define a segment as $\mtraj^{\mathrm{transit}}$ and $\mtraj^{\mathrm{transfer}}$.  

\textit{Object-centric Trajectory: } The mode-augmented trajectory segments can be represented with an object-centric trajectory for transit as $\otraj^{\mathrm{transit}} = \langle \object,\quad [\transform{\object}{\robotstate_t}]_{t=1}^N \rangle$. For transfer, where the relative transform is fixed from the transition $\transform{\object^\times}{\robotstate^\times}$, we can define $\otraj^{\mathrm{transfer}} = \langle \transform{\object^\times}{\robotstate^\times},\quad [\object_t]_{t=1}^N \rangle$. 

\textit{Mode-aware Neighborhood Connections: } Define a object-augmented state as $\modestate = \langle \robotstate, \object \rangle.$ Given two points $( \modestate_i \mode_i )$ and $( \modestate_j \mode_j )$, we can describe some properties of their neighborhoods and neighborhood connections $\traj^{\mathrm{NN}}$. The connection begins at the start query point, $\traj^{\mathrm{NN}}(0) = \modestate_i$. When $\mode_i \neq \mode_j$, we cannot connect the two points with a \textit{simple} continuous eef trajectory $\pi$ that does not contain a mode transition. When $\mode_i = \mode_j$, in $\transit$ an object-centric connection can be defined to move the eef\textit{ with the constraint that the object pose cannot be altered}. The eef needs to be moved to $\traj^{\mathrm{NN}}(1) = (\transform{\object_j}{\robotstate_j} \cdot \object_i,\ \ \object_i )$.  When $\mode_i = \mode_j$, in $\transfer$ an object-centric neighborhood connection alters the object pose but preserves the transition transformation $\transform{\object^\times}{\robotstate^\times}$ to yields $\traj^{\mathrm{NN}}(1) = (\transform{\object^\times}{\robotstate\times} \cdot \object_j,\ \ \object_j )$ as the target of the connection.

\textit{Assumption (Object-centric Task) } We assume that the object centric trajectory is sufficient to describe the underlying task constraint, i.e, transforming the demonstration set to an object-centric form preserves the task information. 

\paragraph{Mode-aware Transformed Imitation: } We can now state the imitation learning objective we study. For a set of object interaction demonstrations, given a query to $\Policy_\demoset$ as $(\robotstate_{\mathrm{query}}, \obs_\mathrm{query}, \mode_\mathrm{query})$, the task constraint trajectory from some $(\robotstate_{\mathrm{near}}, \obs_{\mathrm{near}}, \mode_\mathrm{query})$ which belongs to a demonstration $ \demo_\mathrm{near} \in \demoset$ can be composed using  (i) the connection $\traj^{\mathrm{NN}}$ to join the demonstration, followed by (ii) the object-centric demonstration corresponding to $\demo$ from $(\robotstate_{\mathrm{near}}, \obs_{\mathrm{near}}, \mode_\mathrm{query})$.

The specifics of how the primitives are designed will be presented in the next sections, including the choice of $\demo_\mathrm{near}$, the planning for $\traj^{\mathrm{NN}}$, and closed loop repeated invocation of the policy query.

\section{Method} 
\label{sec:method}
RE4 composes four lightweight steps---\Expansion---inside a receding-horizon loop, with every step conditioned on the current \emph{manipulation mode}. The design rests on two principles: ~(i) \textbf{Relative reframing} expressing a demonstration's actions relative to the manipulated object rather than the world lets a single demonstration cover a continuum of object placements. ~(ii)~\textbf{Manipulation mode} conditioning every step on whether the object is grasped keeps RE4 correct across the discontinuity at contact $\mode\in\{\transit,\transfer\}$
Retrieval and reframing are both expressed relative to the pose of the object being manipulated, $\object \in SE(3)$, in the world frame. We use this object-centric framing around object pose, rather than a
task-defined goal frame, because it is precisely the quantity that relative reframing transports demonstrations into. In the single-object setting we study, the manipulated object is unique, though potentially involving interactions with parts of the scene. The rest of the section explains the steps of the framework.

\textbf{The \acronym Loop.} Algorithm~\ref{alg:re4-rollout} describes RE4 as a receding-horizon loop over its four steps, threading the demonstrations $\demoset$, the current observation $\obs$, and robot state $\robotstate$ through a small set of enabling functions.
\begin{algorithm}[t]
\label{alg:RE4}
\caption{\textsc{RE4 Loop}}\label{alg:re4-rollout}
\begin{algorithmic}[1]
\Require Demonstrations $\mathcal{D}=\{\langle\robotstate_i,\obs_i,\action_i\rangle\}_{i=1}^{N}$ 
\State $\robotstate, \obs \gets \Call{GetObservation}{}()$
\While{$\neg\;\Call{MaxSteps}{}()$ \textbf{and} $\neg\,\Call{Solved}{\robotstate, \obs}$}
    \State $\robotstate, \obs \gets \Call{GetObservation}{}()$ 
    \State $\mode \gets \Call{GetMode}{\robotstate, \obs}$ \Comment{$\mode \in \{\small \textsc{transit},\,\textsc{transfer}\}$}
    \State $\object \gets \Call{GetPose}{\obs}$ 
    \State $\modestate \gets \langle\robotstate, \object\rangle$
    \State $j^\star \gets \Call{Retrieve}{\modestate,\,\mathcal{D},\,\mode}$ 
    \Comment{demo frame lookup}
    \State $\hat{\action}_{1:\horizon},\; \robotstate_{\mathrm{near}} \gets \Call{Reframe}{j^\star,\,\modestate, \mode}$ 
    \State $\hat{\action}_{\mathrm{plan}} \gets \Call{Replan}{\robotstate, \obs,\,\robotstate_{\mathrm{near}}}$ \Comment{EEF to chunk start} 
    \State $\Call{Replay}{\hat{\action}_{\mathrm{plan}}, \hat{\action}_{1:\horizon}, \acthorizon}$ \Comment{execute $\acthorizon + \| \hat{\action}_{\mathrm{plan}} \|$ actions}
\EndWhile
\end{algorithmic}
\end{algorithm}
Each iteration first reads the current observation and state, infers the manipulation mode $\mode$ and the manipulated-object pose $\object$, and forms the query $\modestate = \langle \robotstate, \object \rangle$. \textsc{Retrieve} returns the nearest object-aware demonstration frame $j^\star$; \textsc{Reframe} transports that frame's action chunk into the current object frame and emits an end-effector set-point $\robotstate_{\mathrm{near}}$; \textsc{Replan} drives the end-effector to $\robotstate_{\mathrm{near}}$; and \textsc{Replay} executes the reframed chunk. The loop repeats until task success or the step budget is exhausted. We now present the task-agnostic functionality.

\subsection{Pose Estimation}
\label{sec:pose-estimation}
A key component of RE4 is object pose detection, enabled by \textsc{GetPose} which maps an observation $\obs$ to the world-frame pose $\object$ of the manipulated object. This pose $\pose_\object$ conditions both \textsc{Retrieve} and \textsc{Reframe}. 
\medskip
\subsubsection{Self Supervised Pose Estimator}
\label{sec:self-sup}
For tasks involving rigid attachment to objects (i.e., grasping), we observe that \textsc{GetPose} can be learned in a self-supervised fashion, from the \textit{{demonstration data alone}}; observation data and agent states, under a series of \textit{assumptions}. We identify a proxy for object pose is the pose of the EEF itself. In particular, when the demonstration is in a $\transfer$ state (obj. and EEF are coincident), the pose of the object can be approximated to the pose of the EEF itself. For $\transit$ frames (obj stationary), we can use EEF pose at boundary points between $\transfer$ and $\transit$ modes as the approximate pose of the object itself. This solves the degenerate case where the vision encoder learns only EEF position. This approach requires the assumptions that DOF(eef) = DOF(obj) and there exists a label/proxy for frames where object and eef are coupled/in-contact (e.g., gripper state). 

We use a feed-forward CNN for the self-supervised pose estimator. An ImageNet-pretrained ResNet-18 backbone maps the $84 \times 84$ agentview image (center-cropped to $76$) to a global-pooled $512$-D feature, which a linear layer projects to a $256$-D embedding. A single linear head reads out the pose as a translation $t \in \mathbb{R}^3$ and a $6$-D rotation~\cite{zhou2020rot6d} decoded to $R \in SO(3)$. We train for $20$ epochs with AdamW ($\text{lr} = 3 \times 10^{-4}$, weight decay $10^{-4}$). 
\subsubsection{Pretrained Foundation Model}
\label{sec:foundation-pose}
Foundation models have emerged as strong option for novel object pose estimation. Most notably, NVIDIA's FoundationPose~\cite{Wen_2024_CVPR} exposes a model-free pipeline requiring a set of reference RGB-D images, object segmentation, and object poses which build a neural object representation. 
We collect reference images with a wrist-mounted camera by orbiting the static object. FoundationPose relies on labelled object poses for each reference image to construct its object representation; we instead derive camera extrinsics from the end-effector forward kinematics composed with a calibrated hand–eye transform. 
This yields pose accuracy comparable to COLMAP \cite{schoenberger2016sfm} on well-textured scenes and markedly better accuracy on reflective, poorly-textured objects, at negligible compute cost. This requires our object to be static during collection. We observe this process works sufficiently well, given RE4 requires only relative transforms between object poses rather than a canonical absolute frame (\cref{sec:pose-estimation}).
Following CNOS~\cite{cnos}, we employ a combination of DINOv2 \cite{dinov2} and SAM2 \cite{sam} to perform object segmentation. Requiring a single reference segmentation, we build an embedding using DINOv2 for the target object. We then use SAM2 to generate candidate segmentation's, comparing the DINOv2 embeddings with the known using cosine similarity, to identify target object segmentation. The same process takes place during the start of inference. The segmented object, along with the current RGB-D observation is provided to FoundationPose, returning a pose estimate. For subsequent frames we employ FoundationPose's tracking method, the same process is used to build the RE4 memory.
We observe Any6D \cite{any6d} to be a strong alternative where an orbit cannot be
collected. It requires a single reference RGB-D image with the object masked.
From that one view a 2D-to-3D generative model hallucinates the unseen
geometry---novel views via a multi-view diffusion model, lifted to a mesh by a
feed-forward reconstructor---and the resulting mesh, which is produced only up
to scale, is then aligned to metric size by fitting it against the reference
depth. The representation is passed to FoundationPose for registration and tracking.

\subsection{RE4 Process}

\paragraph{Retrieve}
\label{sec:retrieve}
\textsc{Retrieve} chooses \emph{which} demonstration to follow and \emph{where} to follow
It returns a single source frame $j^\star\in\demo_\mathrm{near}\in\demoset$ through a two-stage lookup, with manipulation mode as a hard filter, ensuring that only frames with matching modes are admissible. Initially, the eef and object distance $\ell_{\modestate}$ is minimized to yield one frame per demo, giving a set of frames $K$;
then the frame with the nearest object pose, $\ell_{\object}$, from $K$ is returned. 

\paragraph{Reframe}
\textsc{Reframe} instantiates the object-centric trajectory $\otraj$ of \cref{sec:foundations} at the current object, producing both the chunk to replay and its entry point. It preserves the object-relative agent pose $\transform{\object}{\robotstate}\triangleq\object^{-1}\robotstate$ as
$
\big(\object^{q}\big)^{-1}\robotstate_{\mathrm{near}}
=\big(\object^{j^\star}\big)^{-1}\robotstate^{j^\star},
$
so the set-point $\robotstate_{\mathrm{near}}=\object^{q}\,\transform{\object^{j^\star}}{\robotstate^{j^\star}}$ is exactly the connection target $\traj^{\mathrm{NN}}(1)$. With the object-pose delta $\delta\triangleq\object^{q}(\object^{j^\star})^{-1}$. In {Transit} ($\otraj^{\transit}$), the object is world-fixed, so $\delta$ carries the demonstrator onto our object: $\robotstate_{\mathrm{near}}=\delta\,\robotstate^{j^\star}$ and $\hat{\action}_k=\delta\,\action^{j^\star}_k$.

For {Transfer} ($\otraj^{\transfer}$), the object is held and the goal
world-fixed, so it is the demonstrator's \emph{object} path that must be
reproduced---through our own grasp, which need not match the
demonstrator's. Requiring
$\hat{\robotstate}_k\big(\transform{\object^{q}}{\robotstate^{q}}\big)^{-1}
=\object^{j^\star}_k$ yields one constant grasp-difference correction
$
C\triangleq\big(\transform{\object^{j^\star}}{\robotstate^{j^\star}}\big)^{-1}
\transform{\object^{q}}{\robotstate^{q}},
$
applied on the \emph{right}: $\hat{\action}_k=\action^{j^\star}_k\,C$ and
$\robotstate_{\mathrm{near}}=\robotstate^{j^\star}C$. When the two grasps agree, $C=\mathbb{I}$, the chunk replays unchanged and $\robotstate_{\mathrm{near}}=\robotstate^{j^\star}$, i.e.\ the connection step still drives the agent onto the demonstrator.

\paragraph{Replan}
\textsc{Replan} realises the mode-aware connection $\traj^{\mathrm{NN}}$ of \cref{sec:foundations}: a path from the current state $\modestate$ ($=\traj^{\mathrm{NN}}(0)$) to the reframed set-point $\robotstate_{\mathrm{near}}$ ($=\traj^{\mathrm{NN}}(1)$), which we solve with standard motion planning. The active mode sets the constraint the connection must respect---in $\transit$ the object pose is held fixed (the eef moves without disturbing the object---implemented using RRT~\cite{lavalle2001randomized} for Push-T state-based); in $\transfer$ the grasp transform $\transform{\object^\times}{\robotstate^\times}$ is preserved. Because \textsc{Retrieve} admits only same-mode frames, the cross-mode case---which
has no simple connection---never arises. The $\transfer$ connection presupposes the object is actually held; a failed grasp violates that invariant and strands the agent in $\transfer$. It is straightforward to detect and handle this failure in execution, bypassing retrieval and reset to a fallback state, here chosen trivially to be initial state.

\paragraph{Replay}
\textsc{Replay} consumes the replan connection as well as the suffix of the retrieved demonstration. At this stage, the executor stitches together the connection plan and a replay horizon from the transformed relative demonstration.

\section{Experiments}

The experiments are reported in this section. All methods in simulation are trained on an RTX 4090, while the real robot experiment used an RTX 5090.
\subsection{Benchmark Tasks}
\label{sec:bench}

\paragraph{\textbf{Simulated}} We benchmark six tasks across three simulator suites and a real-robot. Push-T~\cite{florence2021implicitbehavioralcloning} is a planar ($SE(2)$) task in which an agent guides a 2D \textit{T}-block to a fixed goal pose, containing 206 human expert demonstrations. Robomimic's~\cite{robomimic2021} Lift, Can, and Square and MimicGen's \cite{pmlr-v229-mandlekar23a} Threading tasks.
Each task uses 200 demonstrations.
All tasks entail an object interaction, with Square and Threading in particular requiring fine-grained interactions near the goal.
We test two variants: \emph{(i) state-based}, observing the proprioceptive state $\robotstate$ together with the ground-truth object pose $\object$; and \emph{(ii) image-based}, 
observations include the camera feed overlooking the workspace with $\robotstate$. 
The state-based setting has object pose in observations while image-based settings require pose estimation from observations.

\paragraph{\textbf{Real Robot}} We benchmark a real-robot multi-step \textit{flip the mug} task on a UFactory xArm7. The task involves knocking an upside down mug onto it's side, followed by picking it up and placing upright on the table. We record 20 expert demonstrations for the task. 

\subsection{RE4 Implementation and Ablation}
We instantiate the RE4 framework defined in \cref{sec:method} per task, specializing each module to the respective task.
An ablation is introduced as RE2, lacking \textsc{Reframe} and \textsc{Replan}.
The state-based variant has access to the object pose directly. Image-based benchmarks need a pose estimator module, leading to the designs reported below. 
\paragraph{Push-T} 
For the image-based variant, noted as \textbf{RE4-SS}, using prior work~\cite{wolberg_robust}, we estimate position from the colour mask centroid, and orientation from the rotation-as-shift property of the log-polar transform.
\paragraph{Robomimic/Mimigen} Two pose estimation pipelines are ablated; \textbf{RE4-SS}, a learned encoder (a pretrained ResNet-18 \cite{he2015deepresiduallearningimage} with a linear head) trained under our self-supervised, mode-conditioned objective
following \cref{sec:self-sup}; \textbf{RE4-FP}, an off-the-shelf estimator, employing 
the model-free workflow of 
Foundation Pose~\cite{Wen_2024_CVPR}, SAM2 \cite{sam} and DINOv2 \cite{dinov2} for object pose estimation. Details in \cref{sec:foundation-pose}.
\paragraph{Real Robot} In the real robot setting
we
apply Any6D \cite{any6d} with SAM2 \cite{sam} and DINOv2 \cite{dinov2} to generate the object model, as described in \cref{sec:foundation-pose}. This model is subsequently passed to Foundation Pose \cite{Wen_2024_CVPR} for real-time object pose tracking. 



\begin{table*}[t]
\centering

\begin{subtable}[t]{0.71\textwidth}
\vspace{0pt}
\centering
\setlength{\tabcolsep}{3pt}
\begin{tabular}{l cc cc cc cc c}
\toprule
 & \multicolumn{2}{c}{\includegraphics[width=1.5cm]{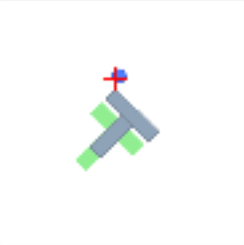}}
 & \multicolumn{2}{c}{\includegraphics[width=1.5cm]{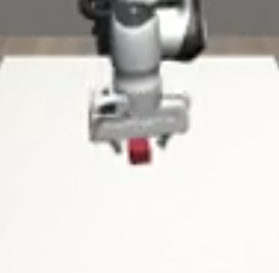}}
 & \multicolumn{2}{c}{\includegraphics[width=1.5cm]{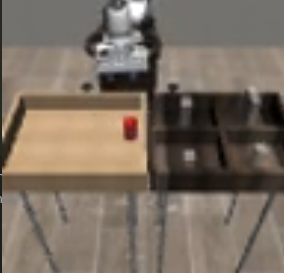}}
 & \multicolumn{2}{c}{\includegraphics[width=1.5cm]{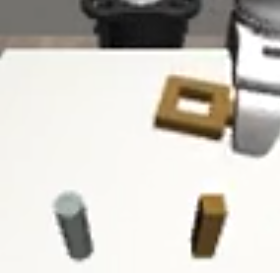}}
 & \includegraphics[width=1.5cm]{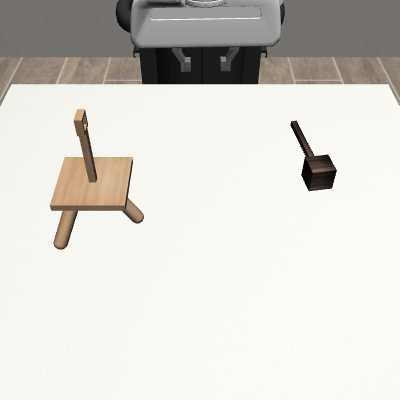} \\
Method 
 & \multicolumn{2}{c}{Push-T \cite{florence2021implicitbehavioralcloning}}
 & \multicolumn{2}{c}{Lift \cite{robomimic2021}}
 & \multicolumn{2}{c}{Can \cite{robomimic2021}}
 & \multicolumn{2}{c}{Square \cite{robomimic2021}}
 & Threading \cite{pmlr-v229-mandlekar23a} \\
\cmidrule(lr){2-3} \cmidrule(lr){4-5} \cmidrule(lr){6-7} \cmidrule(lr){8-9} \cmidrule(lr){10-10}
 & State & Image & State & Image & State & Image & State & Image & Image \\
\midrule
SFP \cite{pmlr-v305-jiang25a}   & 0.859          & --                    & 0.94         & --           & 0.95          & --            & \textbf{0.97} & --            & --   \\
GPI \cite{ICLR2026_bf235a1d}         & 0.912          & 0.624                 & \textbf{1.0} & 0.90         & 0.28          & 0.48          & 0.81          & 0.51          & 0.00 \\
VINN \cite{Pari-RSS-22}                               & --             & --                    & --           & 0.73         & --            & 0.53          & --            & 0.58          & 0.00 \\
DDPM-100 \cite{chi2024diffusionpolicy}                & 0.894          & 0.798                 & \textbf{1.0} & \textbf{1.0} & \textbf{0.98} & 0.97          & 0.91          & 0.84          & 0.75 \\
DDIM-10 \cite{chi2024diffusionpolicy}                 & 0.901          & 0.826                 & \textbf{1.0} & \textbf{1.0} & 0.97          & 0.97          & 0.91          & 0.84          & 0.71 \\
\midrule
RE2 (Ablation)                                         & 0.255          & 0.304                 & 0.69         & 0.73         & 0.29          & 0.65          & 0.25          & 0.34          & 0.11 \\
\textbf{RE4} (State)                                        & {\textbf{0.919}} & -- & {\textbf{1.0}} & -- & {0.95} &           & \textbf{0.97} & --         & -- \\
\textbf{RE4-SS}                                        & -- & {\textbf{0.883}*} & -- & \textbf{1.0} & -- & 0.93          & -- & 0.87          & 0.38 \\
\textbf{RE4-FP}                                        &                                 &                                  &                               & 0.98         &                      & \textbf{0.98} &                                 & \textbf{0.97} & \textbf{0.83} \\
\bottomrule
\end{tabular}
\caption{\textbf{State- and image-based performance.} Push-T reports average maximum coverage; RoboMimic and MimicGen report success rate (100 seeds). 
}
\label{tab:main_benchmark}
\end{subtable}%
\hspace{0.03\textwidth}%
\begin{subtable}[t]{0.25\textwidth}
\vspace{0pt}
\centering
\includegraphics[trim={0.2cm 0.5cm 16.6cm 1cm}, clip, width=0.99\linewidth]{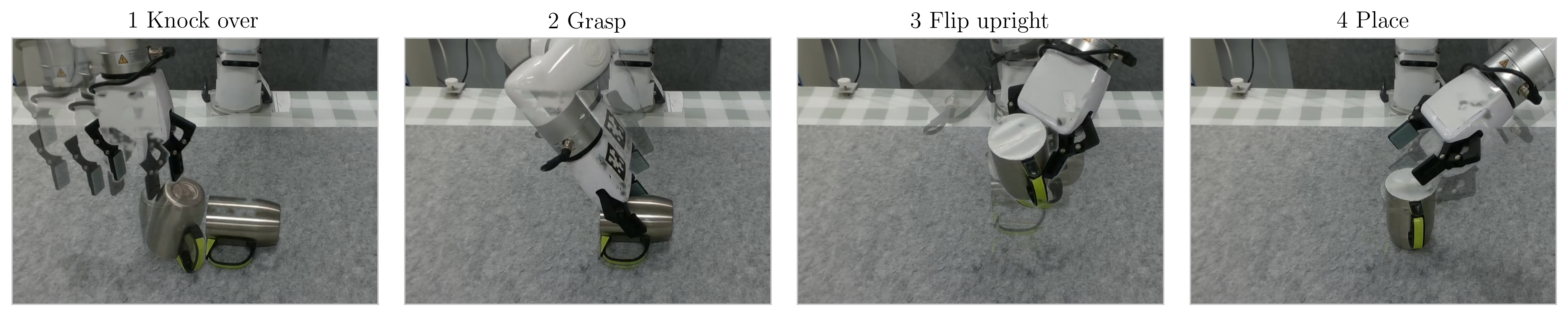} \\[2pt]
\includegraphics[trim={16.8cm 0.5cm 0 1cm}, clip, width=0.99\linewidth]{figures/mug_flip_task.png} \\[4pt]
\setlength{\tabcolsep}{3pt}
\begin{tabular*}{\linewidth}{@{\extracolsep{\fill}} lcc @{}}
\toprule
Method & Success & Steps \\
\midrule
DDPM-100           & 0.60 \tiny{(12/20)}          & $309 \pm 35$ \\
\textbf{RE4-Any6D} & $\mathbf{0.95}$ \tiny{(19/20)} & $\mathbf{211 \pm 15}$ \\
\bottomrule
\end{tabular*}
\caption{\textbf{Real-robot Mug Flip.} Success over 20 trials reported. Average steps reported over successful runs. Task involves toppling the cup, then grasping at an angle, to flip it and placing it upright.
}
\label{tab:mug_flip_combined}
\end{subtable}

\caption{\textbf{Simulated and real-robot benchmark tasks.}}
\vspace{-0.2in}
\label{tab:overall_results}
\end{table*}

\subsection{Baselines and Metrics}
We compare our method against several state of the art approaches, including, Diffusion Policy (DP) \cite{chi2024diffusionpolicy}, Streaming Flow Policy \cite{pmlr-v305-jiang25a}, Geometry aware Imitation Policy \cite{ICLR2026_bf235a1d}. As per previous work \cite{pmlr-v305-jiang25a, ICLR2026_bf235a1d, chi2024diffusionpolicy}, we benchmark the DDPM and DDIM sampling process for each Diffusion Model
at 100 and 10 inference steps respectively \cite{chi2024diffusionpolicy}. These serve as the generative policies. As a lookup-based baseline we use Visual Imitation via Nearest Neighbors \cite{Pari-RSS-22}.  
In the real robot trials, we benchmark against DDPM-100, trained under the same configuration and cropped resolution as Robomimic/Mimicgen per Jiang et. al~\cite{pmlr-v305-jiang25a}.
For Push-T \cite{florence2021implicitbehavioralcloning}, in line with previous work~\cite{chi2024diffusionpolicy}, we measure maximum coverage per episode---where coverage is target area IoU---taking the mean across all seeds. For Robomimic \cite{robomimic2021}, MimicGen \cite{pmlr-v229-mandlekar23a} and the real robot, we report average success.


\subsection{Studies}

In addition to raw performance in the benchmark tasks, we additionally set up extensive studies to test generalization and robustness. 
\textbf{Sparse observation trials} test sample start states for problems till an adversarial set of problems can be selected whose starting state is farthest away from starting states in the demonstration dataset.
\textbf{Sparse demonstration trials} test robustness when trained on only 50 demonstrations. 
\textbf{A visual distractor study} introduces distractor objects (1-3) to the scene which were not present during original demonstrations. 
\textbf{A collision-robustness study} reports the performance when a new geometry in introduced that blocks can task during runtime.

\subsection{Results}
\label{sec:exp-results}

\paragraph{Task Performance}
Table~\ref{tab:main_benchmark} reports performance in the simulated benchmark tasks. 
Both state-based and image-based variants are reported, with empty entries left where inapplicable. Note that SFP has no publicly reported stable training regime for Robomimic/MimicGen and hence was omitted from image-based entries. Overall data indicates that the RE4 variants dominate performance across almost all the tasks, with the exception of RE4-SS on Can and Threading. Notably, RE4 is achieving this performance with lightweight-to-no training. The nearest competitor is DP~\cite{chi2024diffusionpolicy} in these tasks. 
The poor performance of the RE2 ablation demonstrates a clear value of \textsc{Reframe} and \textsc{Replan}, i.e., retrieval alone is not enough. Among the variants of RE4, RE4-FP has more robust performance, motivating the use of general-purpose powerful plug-and-play  pose estimator modules rather that task-specific training of such an estimator. 

On the real-robot experiment, reported in Table~\ref{tab:mug_flip_combined}, RE4 again indicates significant robustness and has significantly better performance compared to DP. The task involves two quite deliberate contacts, one non-prehensile, one prehensile. Qualitatively, DP struggled when the starting location of the cup was not close to one of the demonstrations, while RE4 managed to transform the experience to these settings and finish the task faster as well. The pose estimator was fairly robust through the execution, which helped preserve the performance. The one reported failure resulted from cup toppling after being dropped upright.

\begin{table*}[t!]
\centering
\footnotesize

\begin{subtable}[t]{0.63\textwidth}
\centering
\setlength{\tabcolsep}{2.5pt}
\begin{tabular}{l @{\hspace{8pt}} ccc @{\hspace{12pt}} ccc @{\hspace{12pt}} ccc @{\hspace{12pt}} ccc}
\toprule
 & \multicolumn{3}{c}{\includegraphics[width=1.7cm]{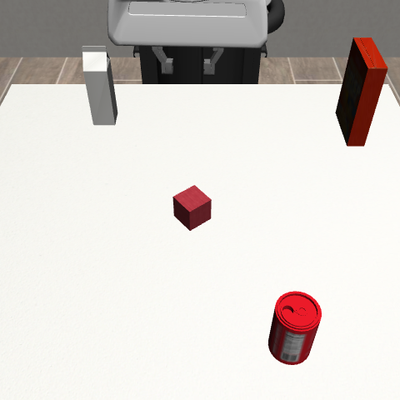}}
 & \multicolumn{3}{c}{\includegraphics[width=1.7cm]{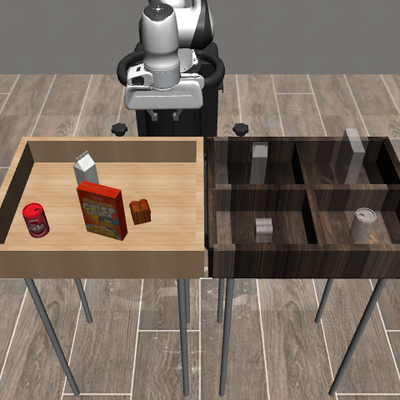}}
 & \multicolumn{3}{c}{\includegraphics[width=1.7cm]{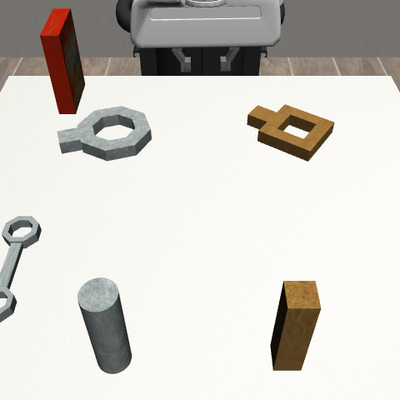}}
 & \multicolumn{3}{c}{\includegraphics[width=1.7cm]{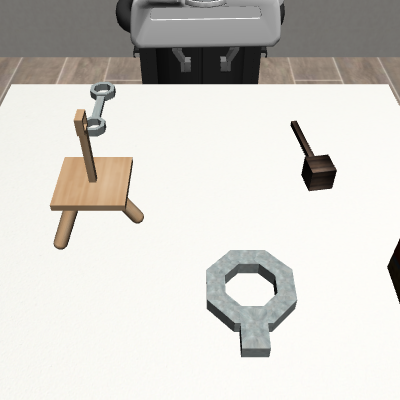}} \\
 & \multicolumn{3}{c}{Lift \cite{robomimic2021}}
 & \multicolumn{3}{c}{Can \cite{robomimic2021}}
 & \multicolumn{3}{c}{Square \cite{robomimic2021}} 
 & \multicolumn{3}{c}{Threading \cite{pmlr-v229-mandlekar23a}} \\
\cmidrule(lr){2-4}\cmidrule(lr){5-7}\cmidrule(lr){8-10}\cmidrule(lr){11-13}
Method & $d_1$ & $d_2$ & $d_3$
       & $d_1$ & $d_2$ & $d_3$
       & $d_1$ & $d_2$ & $d_3$
       & $d_1$ & $d_2$ & $d_3$ \\
\midrule
DDPM-100 \cite{chi2024diffusionpolicy}
  & \textbf{1.00} & \textbf{0.99} & \textbf{0.98}
  & 0.26 & 0.30 & 0.08
  & 0.84 & 0.86 & 0.78 
  & 0.46 & 0.47 & 0.22 \\
\textbf{RE4-SS}
  & 0.65 & 0.52 & 0.33 
  & 0.14 & 0.04 & 0.01 
  & 0.57 & 0.55 & 0.25
  & 0.03 & 0.01 & 0.01 \\
\textbf{RE4-FP}
  & 0.95 & 0.95 & 0.92
  & \textbf{0.99} & \textbf{0.98} & \textbf{0.96}
  & \textbf{0.96} & \textbf{0.97} & \textbf{0.96}
  & \textbf{0.82} & \textbf{0.74} & \textbf{0.76} \\
\bottomrule
\end{tabular}
\caption{Success rates with 1,2, and 3 distractor objects ($d_1\text{--}d_3$, 100 seeds).}
\label{tab:robustness}
\end{subtable}%
\hspace{0.05\textwidth}
\begin{subtable}[t]{0.28\textwidth}
\centering
\setlength{\tabcolsep}{4pt}
\begin{tabular*}{\textwidth}{@{\extracolsep{\fill}} l cc @{}}
\toprule
\multicolumn{3}{c}{\includegraphics[width=2.8cm]{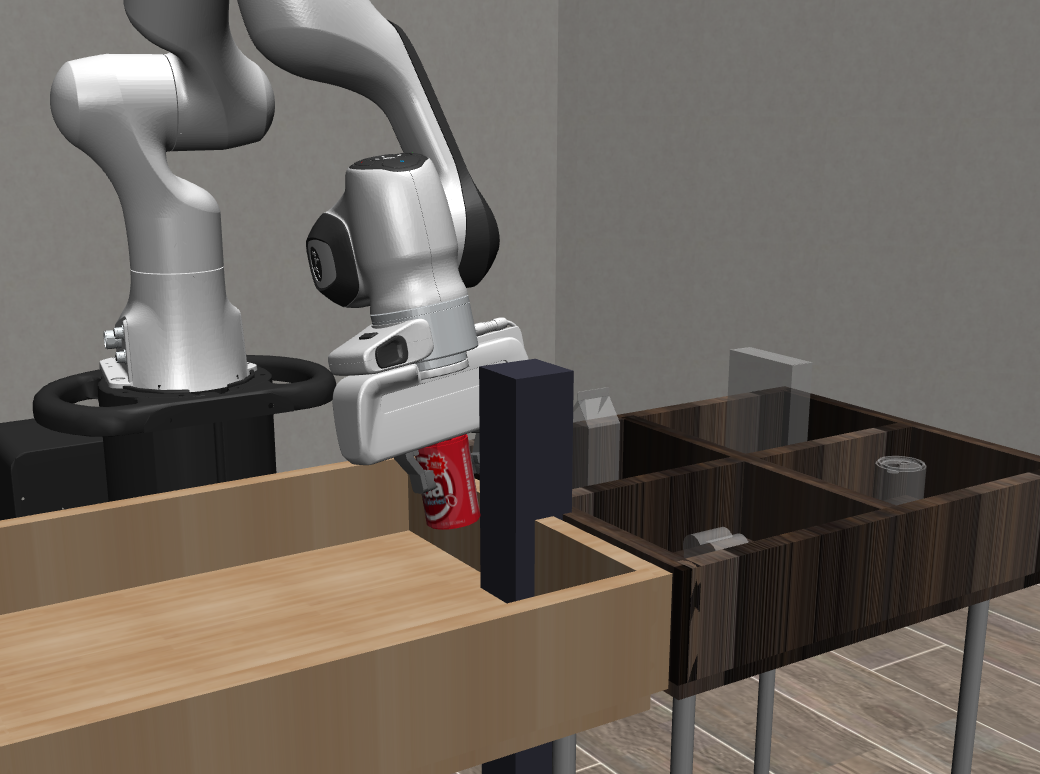}} \\[2pt]
Method & Success
\\
\midrule
DDPM-100 \cite{chi2024diffusionpolicy}  & 0.34    \\
\textbf{RE4-FP} (filter off)             & 0.42     \\
\textbf{RE4-FP} (filter on)              & \textbf{0.74}  \\ 
\bottomrule
\end{tabular*}
\caption{Collision avoidance (50 seeds).}
\label{tab:collision}
\end{subtable}

\caption{\textbf{Robustness to visual distractors and collisions.}}
\label{tab:combined_benchmarks}
\end{table*}

\begin{figure}[t]
\centering
\begin{minipage}[c]{0.58\columnwidth}
  \centering
  \includegraphics[width=\linewidth]{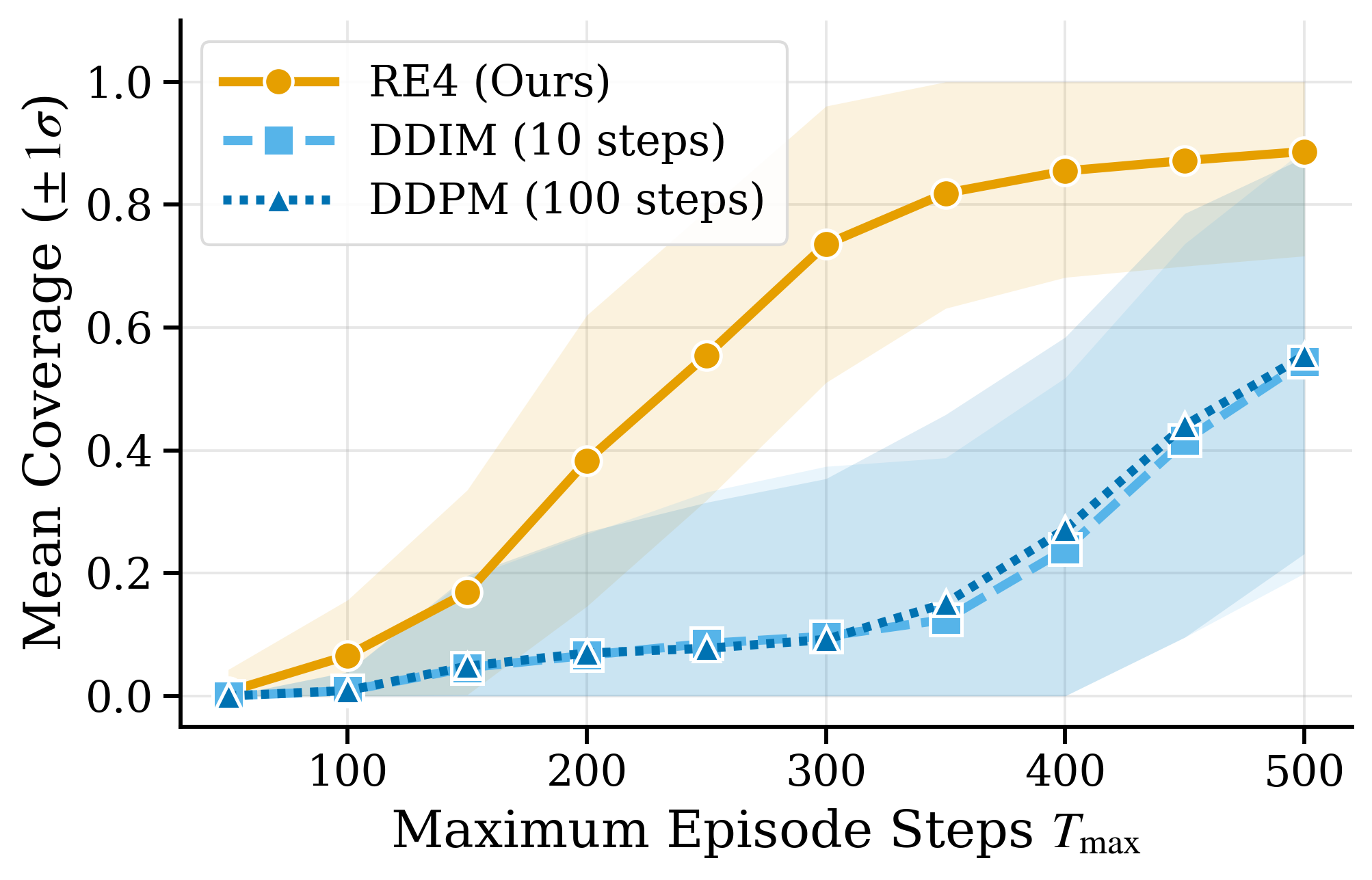}
\end{minipage}%
\hfill
\begin{minipage}[c]{0.4\columnwidth}
  \centering
  \scriptsize
  \setlength{\tabcolsep}{3pt}
  \begin{tabular*}{\linewidth}{@{\extracolsep{\fill}} l cc @{}}
    \toprule
    Method & State & Image \\
    \midrule
    DDPM-100     & 0.531          & 0.609          \\
    DDIM-10      & 0.507          & 0.579          \\
    \textbf{RE4} & \textbf{0.781} & \textbf{0.733} \\
    \bottomrule
  \end{tabular*}
\end{minipage}
\caption{\textbf{Push-T under data sparsity.} 
\textbf{Left:} Mean coverage over\ maximum horizon over $75$ problems sparsely covered by $\demoset$). RE4 is robust while DP performs poorly even when given additional horizon.
\textbf{Right:} Mean coverage when methods are trained with $50$ demos (averaged over $100$ seeds).}
\vspace{-0.2in}
\label{fig:pusht_sparsity_combined}
\end{figure}
\paragraph{Sparse Observations and Demonstrations}
A key hypothesis in this study is the importance of the retrieval process in performance. Fig~\ref{fig:pusht_sparsity_combined} reports a study in the Push-T benchmark. The left image shows performance over $75$ adversarial problems, selected from starting states that are farthest away from demonstrated problems, RE4 has shown significant performance benefits, which manage to keep improving as the execution horizon is allowed to increase. 
\begin{wraptable}{r}{0.56\columnwidth}
  \vspace{-1.0em}
  \centering
  \footnotesize
  \setlength{\tabcolsep}{2.5pt} 
  \caption{Success in the 75 sparse problems on image-tasks.}
  \label{tab:low-density}
  \begin{tabular}{l cccc}
    \toprule
    Method & Lift & Can & Square & Thread. \\
    \midrule
    DDIM-10         & 0.81          & 0.92          & 0.09          & 0.57          \\
    DDPM-100        & 0.84          & 0.92          & 0.12          & 0.59          \\
    \textbf{RE4-SS} & \textbf{0.97} & \textbf{0.95} & 0.63          & 0.20          \\
    \textbf{RE4-FP} & \textbf{0.97} & 0.85          & \textbf{0.97} & \textbf{0.72} \\
    \bottomrule
  \end{tabular}
  \vspace{-0.8em}
\end{wraptable} 
The table on the right reports results from both state and image-based variants when trained only on $50$ demonstrations, which is a quarter of the original set. Here again, DP plummets in performance, in contrast to RE4. 

Table~\ref{tab:low-density} reports the adversarial problem selection for the Robomimic/Mimicgen image-based benchmarks. The starting states are selected to be $75$ problems that are most distinct from the demonstration dataset. DP here still struggles but to different extents. We observe that the problems differ in the nature of (the default) sampling domain for starting states. 
This presumably stymies DP in its generalization, i.e., whether it has observed similar problems in the dataset. RE4 is still better or competitive than DP.

\paragraph{Visual Distractors} 
In each of the image-based Robomimic/MimicGen benchmark tasks, we add 1,2, and 3 distractor objects, reported as $d_1,d_2,d_3$ (Table~\ref{tab:robustness}). This study reveals the sensitivity of the performance to the visual cues present in the image. We observe that using FoundationPose in RE4-FP maintains the best or competitive performance across all four tasks. Interestingly the self-supervised version RE4-SS shows severe brittleness. This is explained by the likely overfitting to the demonstration dataset and the corresponding sensitivity to additional objects. 
DP appears robust only in Lift and Square because the (default) starting locations are very similar, i.e., it can ignore visual cues and on position cues alone in this task. 
FoundationPose on the other hand, by virtue of its object awareness, succeeds mostly to keep track of the task-relevant object.

\paragraph{Collision-robustness in Environment Changes} 
In Table~\ref{tab:collision} we report a proof-of-concept test for how the RE4 design is seamlessly amenable to interpretable downstream filtering. In the visual distractors, the object locations may confuse the detector or learning but do not add any significant geometric complexity. In contrast the collision study alters the environment with a pillar-like object in the middle of the workspace. The demonstrations are then filtered and only the ones that are collision-free are preserved for RE4 lookup. The intuition is that, even this relative simple strategy can eliminate obviously or likely unsafe demonstrations. In contrast, DP and the unfiltered RE4 degrade precipitously. 

\section{Limitations}
Multi-object manipulation settings need enumeration and bookkeeping of multiple or sequential manipulation modes which have not been currently tested. This again both a limitation and an opportunity to use the mode-aware reasoning to plug in multimodal rollouts in multi-object settings.  Differing embodiments remain untested, though expected to be amenable to estimated affordance transformations. The current scope limits focus to contact-based affordances, while richer affordances, articulated or deformable interactions will need future inspection. VLAs present a compelling paradigm of scale that foregoes interpretability. The retrieval-based method, while not immediately applicable to the scale of the foundational datasets, poses a line of inquiry towards a future probe into VLA interpretability. 

\section{Discussion}
In this work we have presented several simple first principle ideas borrowed from theories of manipulation to reconstruct a modular framework composed of four key steps of retrieve, reframe, replan, and replay. 
Pose estimation sits at the heart of RE4, conditioning the retrieval metric $d_{\modestate}$ and the \textsc{Reframe} transform, so 
we note this work as a strong motivation
to identify and design the critical building blocks, like pose estimators within simple interpretable frameworks.
Results indicate promising performance in imitation learning benchmarks that outperform baselines, while additionally expressing interpretability (from our retrieval), lightweight training (only image-based pose estimation from demonstration), and robustness in sparse, distractor, and collision filter studies. Future work will explore more integrated combinations of learned modules, topological properties of demonstrations, and planning (motion and multimodal) to compose hybrid extensible architectures that may balance the strengths of long-horizon planning, search, and imitation. The opportunities for exact interpretability allow explaining demonstrations and their contribution to task performance. This promises ushering in performant, robust, intepretable paradigms in learning object interaction tasks.

\appendices
\input{appendix}

\bibliographystyle{IEEEtran}
\bibliography{preprint}  


\end{document}

%% file: notations.tex
\newcommand{\acronym}{RE4\xspace}
\newcommand{\Expansion}{Retrieve, Reframe, Replan, Replay\xspace}
\newcommand{\expansion}{retrieve, reframe, replan, and replay\xspace}
\newcommand{\taskspace}{\mathcal{X}}
\newcommand{\robotstate}{x}
\newcommand{\modes}{\mathcal{M}}
\newcommand{\mode}{m}
\newcommand{\transit}{\modes_{\mathrm{transit}}}
\newcommand{\transfer}{\modes_{\mathrm{transfer}}}
\newcommand{\obs}{\mathcal{O}}
\newcommand{\object}{o}
\newcommand{\eef}{r_{\mathrm{ee}}}
\newcommand{\tf}{\mathcal{T}}
\newcommand{\horizon}{h}
\newcommand{\acthorizon}{n}
\newcommand{\demoset}{\mathcal{D}}
\newcommand{\demo}{d}
\newcommand{\action}{a}
\newcommand{\pose}{p}
\newcommand{\modestate}{q}
\newcommand{\traj}{\pi}
\newcommand{\poses}{\mathcal{P}}
\newcommand{\Policy}{\Pi}
\newcommand{\transform}[2]{\mathtt{Tr}_{#1 \rightarrow #2}}
\newcommand{\mtraj}{\hat{\traj}}
\newcommand{\otraj}{\hat{\traj}_\object}
\newcommand{\transition}{\modes^\times}

%% file: appendix.tex


\section{RE4 Algorithms}

\begin{algorithm}[h]
\caption{\textsc{Reframe}}\label{alg:re4-reframe}
\begin{algorithmic}[1]
\Require frame $j^\star$ with $(\object^{j^\star},\,\robotstate^{j^\star},\,\action^{j^\star}_{1:\horizon})$; query $\modestate$ with $(\object^{q},\,\robotstate^{q})$; mode $\mode$
\State $\delta \gets \object^{q}\,(\object^{j^\star})^{-1}$ \Comment{object-pose delta: demo $\rightarrow$ query}
\If{$\mode = \transit$} \Comment{object fixed in world; transport demo into query's object frame}
    \State $\robotstate_{\mathrm{near}} \gets \delta \cdot \robotstate^{j^\star}$
    \State $\hat{\action}_k \gets \delta \cdot \action^{j^\star}_k,\quad k = 1,\dots,\horizon$
\Else \Comment{$\mode = \transfer$; object moves with agent, world-fixed goal}
    \State $\robotstate_{\mathrm{near}} \gets \delta^{-1} \cdot \robotstate^{q}$ \Comment{express current agent in demo's object frame}
    \State $\hat{\action}_{1:\horizon} \gets \action^{j^\star}_{1:\horizon}$ 
\EndIf
\State \Return $\hat{\action}_{1:\horizon},\; \robotstate_{\mathrm{near}}$\Comment{return actions and target agent position}
\end{algorithmic}
\end{algorithm}

\section{Implementation Details}
\label{sec:impl}
The implementation of RE4 and all benchmarked methods, including training and inference. 

\subsection{RE4}
\label{sec:RE4-implementation}

\subsubsection{\textbf{Robomimic}}
Below we outline the Robomimic \cite{robomimic2021} instantiation of RE4 components.

\paragraph{Manipulation Mode}
We define manipulation mode $\transfer$ when the last action's gripper state is set, and $\transit$ otherwise. This mode also drives the case selection in \cref{eq:pose-abs,eq:pose-rel,eq:pose-cross}.

\paragraph{Retrieve}
We define a scale-normalised Euclidean translation term and a geodesic $SO(3)$ rotation term for both $\object_\pose$ and $\robotstate_\pose$. The full-state distance $d_{\modestate}$ is a weighted sum of the target-pose distance and the object-frame end-effector distance, so the first stage matches both the target's pose and the gripper's pose relative to it; the across-demo stage then scores on $d_{\pose}$ alone.

\paragraph{Replan}
\textsc{Reframe} produces a retrieved action chunk whose first waypoint sits at the target start pose $\robotstate_{near}$, but the agent's current end-effector is generally elsewhere. Before replaying the chunk we \emph{bridge}: a short closed-loop phase that servos the end-effector from its current pose to $\robotstate_{near}$ using the same absolute OSC controller as replay. Each step emits an absolute pose setpoint toward $\robotstate_{near}$, and the bridge terminates once the end-effector is within a position/orientation tolerance.

\paragraph{Failure Identification}
To enable \textit{reset} within \textsc{Replan}, we define \textsc{failure} as a finger-joint separation $\lVert q_{\text{finger}_l}-q_{\text{finger}_r} \rVert < 0.01$ in $\transfer$ mode, a proxy for an unsuccessful grasp.
\medbreak

\subsubsection{\textbf{Push-T}}
Unlike the prehensile Robomimic tasks, where the gripper state cleanly partitions the rollout into $\transit$ and $\transfer$ segments, Push-T \cite{florence2021implicitbehavioralcloning} is a planar, non-prehensile pushing task: the agent never grasps the T-block, so every frame belongs to $\transit$. This collapses the mode filter, restricts \textsc{Reframe} to a single SE(2) transport, and lets us recover the block pose in closed form rather than from a learned encoder.

\paragraph{Pose Estimation}
We estimate the T-block pose in closed form, with no learned network. The block is segmented by colour thresholding the RGB frame and keeping the largest connected component; its pixel centroid maps to a world position through a fixed pixel-to-world affine calibration. Orientation exploits the rotation-as-shift property of the log-polar transform~\cite{wolberg_robust}: we resample the mask in log-polar coordinates about its centroid, so a rotation of the block becomes a circular translation along the angular axis. Cross-correlating this map against a single anchor frame via the FFT, the rotation is read off directly as the angular shift at the correlation peak. The only ground-truth consumed is the task-defined goal pose, used once to calibrate the anchor; every per-frame estimate is otherwise derived purely from the image.

\paragraph{Retrieve}
For Push-T, $d_{\modestate}$ is an $L_2$ distance over the min--max-normalised
5-D state $(\robotstate, \pose_\object)$ (agent position + block pose); the
across-demo $d_{\object}$ restricts to the 3-D block sub-vector. An
$\varepsilon$-tolerance on $d_{\object}$ filters per-episode candidates
before the stage-1 argmin.

\paragraph{Replan}
\textsc{Replan} for Push-T solves a planar collision-free $\robotstate \rightarrow \robotstate_{\mathrm{near}}$ problem with the T-block treated as a static obstacle (the transit constraint). We use a \emph{Rapidly-exploring Random Tree} (RRT) \cite{lavalle2001randomized}: a sampling-based planner that incrementally grows a tree of obstacle-free configurations rooted at the current pusher position, repeatedly extending the nearest tree node by a bounded step toward a uniformly sampled point (with goal-bias to encourage convergence). The result is a polyline from $\robotstate$ to $\robotstate_{\mathrm{near}}$ that the executor consumes as a sequence of absolute waypoints. We use a $15$ world-unit step, a $15$ world-unit pusher keepout radius around the estimated T pose, and a $1000$-iteration budget. When the direct segment between $\robotstate$ and $\robotstate_{\mathrm{near}}$ is itself collision-free we short-circuit the search and execute that segment directly.
\medbreak
\subsubsection{\textbf{Real Robot}}

\begin{figure*}
    \centering
    \includegraphics[width=1.0\linewidth]{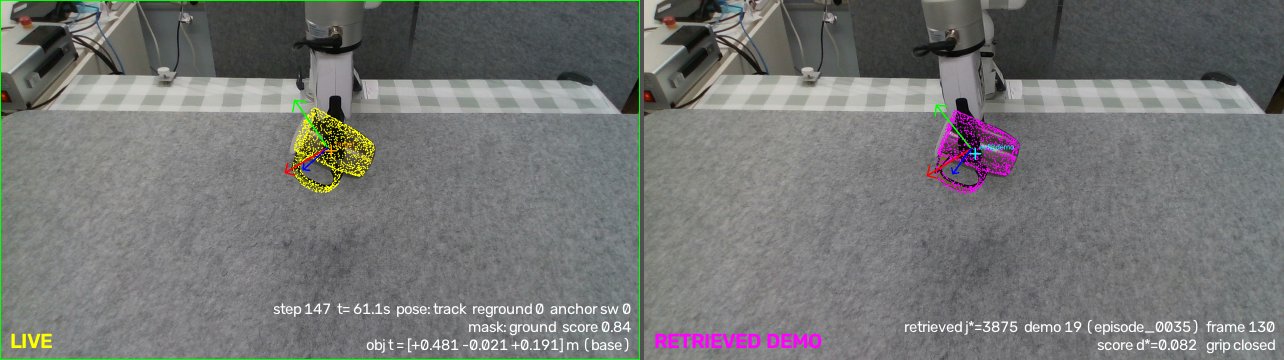}
    \caption{Real-time depiction of estimated mug pose alongside retrieved demo}
    \label{fig:mug-frame}
\end{figure*}

\paragraph{Platform}
A UFactory xArm with a binary parallel gripper, and a single Intel RealSense
mounted in a fixed third-person pose. Colour and depth are streamed at
$1280\times720$ and $30$\,Hz, with depth aligned to the colour frame so that a
single intrinsic $K$ describes both. Because the camera does not move, its
extrinsic $T_{\text{base}}^{\text{cam}}$ is a single constant, recovered once by
eye-to-hand calibration (ArUco marker on the end effector, Tsai--Lenz
$AX=XB$). 

\paragraph{Object model from one reference frame}
The object model is built from a single RGB-D frame using Any6D \cite{any6d}. A coarse box or click prompt seeds
SAM~2 \cite{sam}, which returns the object mask; the masked crop is centred and passed to a view-conditioned diffusion model
(zero123plus \cite{shi2023zero123plus}) to hallucinate novel views, and InstantMesh \cite{xu2024instantmesh} lifts those views to a mesh. That
generative stack recovers \emph{shape} but not \emph{scale}, so the mesh is then rescaled metrically by \texttt{register\_any6d}, which fits the mesh's oriented bounding box against the real depth cloud, searches the extent-ratio permutations, and scores the candidates with FoundationPose's \cite{Wen_2024_CVPR} own refiner and scorer. The complete supervision consumed by this stage is \textbf{one mouse click}.

\paragraph{Pose estimation at rollout}
Segmentation at rollout follows the CNOS recipe \cite{cnos}: SAM~2 emits class-agnostic proposals, each is embedded with DINOv2
\cite{dinov2, sam}, and the proposal whose embedding is most similar to a bank of $8$ onboarding reference crops of the object is taken as the mask. FoundationPose \texttt{register}s once against that mask and \texttt{track\_one}s thereafter at the camera's $30$\,Hz, so the per-step cost is a tracking update rather than the $252$-hypothesis registration sweep.
\cref{fig:mug-frame} shows the estimate live alongside the demonstration frame retrieval selected and that frame's own recorded pose.

\paragraph{Demonstrations}
$20$ demonstrations were collected by joystick teleoperation and logged at $10$\,Hz as RGB, aligned depth, and a $10$-D row
$[\,xyz \in \mathbb{R}^3,\ \text{rot6d} \in \mathbb{R}^6,\ \text{grip} \in \{\pm 1\}\,]$ of absolute end-effector targets---the same layout RE4's memory consumes, so no conversion stands between collection and retrieval.

\paragraph{Algorithmic recipe}
Identical to the Robomimic/MimicGen \cite{robomimic2021, pmlr-v229-mandlekar23a} instantiation of \cref{sec:RE4-implementation}, the same mode rule, the same two-stage retrieval,
the same tolerance-terminated bridge, and the same $(\horizon, T_a) = (24, 8)$ chunk schedule, \textbf{failure identification} is ommitted. Success likewise has no reward signal to read and is instead scored as a verified non-empty grasp followed by a lift of at
least $90$\,mm.

\paragraph{Baseline}
Diffusion Policy is trained on the identical $20$ demonstrations with the image recipe of \cref{sec:impl} unchanged---$84\times84$ RGB, no depth, the same $(T_o, T_p, T_a) = (2,16,8)$ schedule---so the comparison isolates the representation rather than the data or the action interface. Results are in \cref{tab:mug_flip_combined}.

\subsection{\textbf{Diffusion Policy}}
Consistent with reported benchmarking, we train image based diffusion checkpoints for 3000 epochs \& state based for 4500 \cite{chi2024diffusionpolicy}. Training scripts from official repository are replicated.

Both variants use the paper's \cite{chi2024diffusionpolicy} chunk schedule: observation horizon $T_o=2$, prediction horizon $T_p=16$, action horizon $T_a=8$. The noise schedule is DDPMScheduler with a squared-cosine $\beta$ and $\varepsilon$-prediction (clipped samples); the 10-step DDIM is constructed from the trained 100-step DDPM scheduler at inference. State variants regress 2-D (Push-T) or 7-D (Robomimic) actions from the flat state vector; image variants encode each camera with ResNet-18 (BN $\rightarrow$ GN) + SpatialSoftmax ($K=32$) and concatenate with the proprioceptive state. The denoiser is ConditionalUnet1D with sinusoidal time embedding (scale 1). Training: AdamW ($\text{lr}=10^{-4}$, weight decay $10^{-6}$), cosine LR with 500-step warm-up, EMA power $0.75$.

\subsection{\textbf{SFP}}
The public reference implementation \cite{pmlr-v305-jiang25a} publishes a sole state-based Push-T
variant (and no Robomimic variant). We attempt to faithfully replicate the robomimic implementation, keeping the $(T_o, T_p, T_a) = (2,16,8)$ chunk schedule, sampler settings, and streaming dynamics-coefficient at the repository's defaults \cite{chi2024diffusionpolicy, pmlr-v305-jiang25a}. We use the configuration of $1000$ epochs at a batch size of $1024$, AdamW ($\text{lr}=1\mathrm{e}{-4}$, weight decay $1\mathrm{e}{-6}$), a cosine learning-rate schedule with $500$ linear warmup steps, and an exponential moving average of the weights ($\text{power}=0.75$) used for inference. Both the deterministic and stochastic variants use $\sigma=0.1$ (with $\sigma_0=\sigma_1=0.1$ in the stochastic case). 

\subsection{\textbf{GPI}}
We use GPI's \cite{ICLR2026_bf235a1d} default hyperparameters across all experiments: $K{=}3$ nearest neighbours, $\sigma{=}0.01$ Gaussian noise on the normalised query, and softmax temperature inherited from the official repository \cite{ICLR2026_bf235a1d}. The action horizon is $H{=}1$ for state inputs (reactive, paper-recommended) and $H{=}4$ for vision inputs (matches the repository's vision policy). For Robomimic, only the $xyz$ channel is blended by the GPI flow; the rotation and gripper channels are copied from the top-1 neighbour's recorded action, since a weighted sum of rotvecs is not a rotation. The vision encoder is a ResNet-18 (BN$\rightarrow$GN) trained from scratch with MSE supervision on the ground-truth object-state vector. For the state-based Robomimic port, we additionally apply the PushT-style \emph{recent-state suppression} (window $25$) used by GPI's official StateGPIPolicy---without it, the flow stalls at a pre-grasp fixed point on Can/Square and the gripper never closes \cite{ICLR2026_bf235a1d}.

\subsection{\textbf{VINN}}
VINN \cite{Pari-RSS-22} is a self-supervised encoder is trained without action labels, where the policy is a non-parametric nearest-neighbour lookup over the demonstrations in that embedding space. We replicate the
reference implementation.

\paragraph{Representation}
BYOL \cite{byol} over every demonstration frame
of the task's own dataset, with no action supervision. The encoder is an
ImageNet-pretrained ResNet-50 and the augmentation stack is the repository's
verbatim: \texttt{RandomResizedCrop} (scale $0.6$--$1.0$), \texttt{ColorJitter}
$(.8,.8,.8,.2)$ at $p{=}0.3$, \texttt{RandomGrayscale} $p{=}0.2$,
\texttt{GaussianBlur} $p{=}0.2$, then ImageNet normalisation. Training uses Adam
($\text{lr}=3\times10^{-4}$), batch size $32$, $100$ epochs, at $224\times224$.

\paragraph{Memory}
Each frame is embedded once with the augmentation disabled (deterministic resize
and normalise only) and its $2048$-D \texttt{avgpool} activation is stored
alongside the action recorded at that frame. Embeddings are read from the same
layer BYOL optimised through, so the bank cannot diverge from the trained
representation.

\paragraph{Inference}
At each step the current observation is embedded through the same deterministic
transform, the $k$ nearest bank entries are found by Euclidean distance, and the
action is the softmax-weighted mean of their recorded actions,
$\action = \sum_i w_i \action_i$ with $w = \mathrm{softmax}(-d)$ over the raw
distances (no temperature)---VINN's rule unchanged. We use $k{=}1$ on all
tasks, and VINN emits \textbf{one action per step} with no chunking, unlike the
diffusion baselines.

Structurally this is the closest baseline to RE4, both act by retrieving from
the demonstration set rather than by regressing a policy. They differ in
\emph{what} the retrieval key is---a learned image embedding for VINN, an
estimated object pose for RE4---and in whether the retrieved demonstration is
transformed before use, which VINN does not do.

\subsection{\textbf{Training time}}
All checkpoints were trained on a single NVIDIA GeForce RTX 4090 (24\,GB) GPU,
provisioned through RunPod. Per-model training times are reported in
\cref{tab:time}.
\begin{table*}[t]
\centering
\footnotesize
\setlength{\tabcolsep}{4pt}
\caption{\textbf{Training time (minutes).} For each task, the left column
is state-based and the right is image-based. DDPM and DDIM share a trained model, so
their training times are identical.}
\label{tab:time}
\begin{tabular}{l cc cc cc cc cc}
\toprule
 & \multicolumn{2}{c}{Push-T} & \multicolumn{2}{c}{Lift} & \multicolumn{2}{c}{Can} & \multicolumn{2}{c}{Square} & \multicolumn{2}{c}{Threading} \\
\cmidrule(lr){2-3}\cmidrule(lr){4-5}\cmidrule(lr){6-7}\cmidrule(lr){8-9}\cmidrule(lr){10-11}
Method & State & Image & State & Image & State & Image & State & Image \\
\midrule
Diffusion (DDPM/DDIM) & 414m & 489m & 89m & 291m & 214m & 413m & 221m & 434m & -- & 1354m \\
SFP                   & 9m   & -- & 10m  & --  & 25m  & --  & 32m  & -- & -- & -- \\
GPI                   & 0m   & 31m   & 0m   & 6m   & 0m   & 14m  & 0   & 19m & -- & 21m \\
VINN                   & --   & --   & -- & 96m   & --   & 154m  & --   & 220m & -- & 317m \\
RE4 Self Sup          & 0m   & 0m   & 0m   & 3m   & 0m   & 8m   & 0m   & 20m & -- & 32m \\
\bottomrule
\end{tabular}
\end{table*}

\section{Expert Data}
\label{sec:data}
Single human expert data was used for Push-T \cite{florence2021implicitbehavioralcloning} and Robomimic \cite{chi2024diffusionpolicy} tasks, sourced from \url{diffusion-policy.cs.columbia.edu/data/training/}. Each Robomimic task (lift, can, square) contains 200 expert demonstrations, while Push-T contains 206 demonstrations. We use all demonstrations for checkpoints training and rollout memory across all benchmarks---excluding the sparse demonstration experiment \cref{fig:pusht_sparsity_combined}, where only the first 50 for Push-T are used. For Mimigen's \cite{pmlr-v229-mandlekar23a} threading, we use the D0 variant, sampling 200 demonstrations from the set of 1000 synthetic available online. 

\section{\textbf{Self-Supervised Pose Estimation}}
\label{sec:pose-appendix}
\cref{sec:pose-estimation} describes the intuition behind the self-supervised estimator $f_\theta : \obs \mapsto \hat{\pose}_\object$: the end-effector pose serves as a proxy for the object pose in $\transfer$ mode, and the end-effector pose at the $\transit\!\rightarrow\!\transfer$ boundary serves as the proxy in $\transit$ mode. Here we formalise the training objectives, which are conditioned on manipulation mode and require no object model or privileged state.

Let $\poses \supset \taskspace$ be the space of poses (nominally SE(3)) equipped with a metric $\ell: \poses\times\poses \rightarrow \mathbb{R}_{\ge 0}$, and let $\Delta(\pose, \pose')$ denote the relative pose between $\pose, \pose' \in \poses$.

\paragraph{Anchors}
The mode discontinuity yields, for each demonstration $\demo$, a grasp frame $g(\demo) = \min\{\, i \in \demoset_\demo : \mode_i = \transfer \,\}$ at which the end-effector coincides with a graspable object pose. We define the demonstration anchor $\pose^\star_\demo \triangleq \eef^{\,g(\demo)}$. Anchoring pre-grasp predictions to $\pose^\star_\demo$, rather than to the current end-effector, is critical in disambiguating the regression target from the end-effector's own position.

\paragraph{Absolute regression.}
Each frame regresses onto an absolute target: the coincident end-effector pose when grasped ($\transfer$), and the demonstration anchor otherwise ($\transit$):
\begin{equation}
\mathcal{L}_{\mathrm{abs}}(\theta)
= \sum_{\demo} \sum_{i \in \demoset_\demo}
  \ell\!\big(f_\theta(\obs_i),\; y_i\big),
y_i =
\begin{cases}
\eef^{\,i}        & \mode_i = \transfer,\\[2pt]
\pose^\star_\demo & \mode_i = \transit.
\end{cases}
\label{eq:pose-abs}
\end{equation}

\paragraph{Same-demonstration relative consistency.}
For frame pairs within a demonstration, the predicted relative pose should match that of the end-effector when grasped, and vanish otherwise:
\begin{equation}
\begin{aligned}
\mathcal{L}_{\mathrm{rel}}(\theta)
= \sum_{\demo}\Bigg[\;
    &\sum_{\substack{(i,j)\in\demoset_\demo^2 \\ \transfer}}
      \ell\big(\Delta(f_\theta(\obs_i), f_\theta(\obs_j)),\; \Delta(\eef^{\,i}, \eef^{\,j})\big) \\
    &+ \sum_{\substack{(i,j)\in\demoset_\demo^2 \\ \transit}}
      \ell\big(\Delta(f_\theta(\obs_i), f_\theta(\obs_j)),\; \mathbb{I}\big)
  \;\Bigg],
\end{aligned}
\label{eq:pose-rel}
\end{equation}
where $\mathbb{I}$ is the identity transform. The $\transit$ term encodes a static object while the gripper approaches; the $\transfer$ term a rigid grasp that moves the object with the gripper.

\paragraph{Cross-demonstration consistency.}
Pre-grasp predictions from demonstrations whose anchors coincide (within an anchor radius $\rho$) are tied together, yielding a demonstration-invariant object frame for cross-demonstration retrieval and reframing:
\begin{equation}
\mathcal{L}_{\mathrm{cross}}(\theta)
= \sum_{\substack{(\demo,\demo'),\; i\in\transit(\demo),\; i'\in\transit(\demo') \\
    \lVert \pose^\star_\demo - \pose^\star_{\demo'} \rVert < \rho}}
  \ell\!\big(f_\theta(\obs_i),\; f_\theta(\obs_{i'})\big).
\label{eq:pose-cross}
\end{equation}

\paragraph{Combined objective and training.}
The estimator minimises
$\mathcal{L} = \lambda_{\mathrm{abs}}\mathcal{L}_{\mathrm{abs}}
          + \lambda_{\mathrm{rel}}\mathcal{L}_{\mathrm{rel}}
          + \lambda_{\mathrm{cross}}\mathcal{L}_{\mathrm{cross}}$,
which suffices to recover the relative transform between object poses $\object^i, \object^j$ required by \textsc{Retrieve} and \textsc{Reframe}. In practice we use $\lambda_{\mathrm{abs}}{=}1$ (split $0.3/0.7$ grasp/pre-grasp), $\lambda_{\mathrm{rel}}{=}0.8$ (split $0.5/0.3$ zero-/grasp-delta), and $\lambda_{\mathrm{cross}}{=}0.5$; translation and rotation terms in $\ell$ are weighted $10{:}1$ (m$^2$ : rad). Architecture and optimisation settings follow \cref{sec:pose-estimation}: an ImageNet-pretrained ResNet-18 with a $256$-D embedding and a $3{+}6$-D pose head, trained for $20$ epochs with AdamW ($\text{lr} = 3\times10^{-4}$, weight decay $10^{-4}$). Supervision comes entirely from recorded end-effector trajectories and gripper state; no ground-truth object poses are used.

\section{Sparse Data Benchmarks}
  \label{sec:sparsecol}

  Imitation-learning policies are well known to degrade once they encounter states far 
  from the demonstrated distribution, where errors compound 
  \cite{ross2011reductionimitationlearningstructured}; retrieval-based
  methods are especially exposed, since by construction they reuse the
  nearest demonstration frame and have nothing fall back on when no
  demonstration is nearby \cite{sridhar2024memoryconsistentneuralnetworksimitation}. 
  
  \subsection{\textbf{Sparse Observation Experiment}}
  
To make this regime measurable, we evaluate Push-T on \emph{adversarially-sparse} initial conditions: scenes whose object pose maximises the distance to every demonstration's object pose.

  Formally, let $s_\object = \pose_\object \in SE(2)$ denote the object
  pose and $\demoset$ the demonstration support. The coverage of a
  candidate scene is the distance to its nearest demonstration frame:
  \begin{equation}
    \rho(s_\object) \;=\; \min_{i \in \demoset}
                            \lVert s_\object - s_\object^i \rVert,
  \end{equation}
  and the $K$ probed scenes are the $K$ poses with the largest $\rho$,
  subject to a minimum-separation constraint that prevents the $K$ picks
  from clustering in a single tail:
  \begin{equation}
    \{s^\star_k\}_{k=1}^{K}
    \;=\;
    \text{top-}K\ \arg\max\, \rho(s_\object).
  \end{equation}
  
  We find these by sweeping a $50^3$ grid over the min-max-normalised
  $(x,y,\theta)$ object-pose space (computed from $\demoset$, padded by 2\,cm in
  position and $0.15$\,rad in yaw to admit out-of-support gaps), and greedily
  selecting the $K$ candidates with the largest $\rho$ that respect the
  separation constraint. Sparsity is measured \emph{over the object alone}:
  for each $s^\star_k$ we then draw $J$ pusher positions uniformly across the
  workspace, rejecting samples whose disc overlaps the T-block keepout. The
  resulting $K \cdot J$ scenes isolate the regime where the
  \emph{scene} is hard to retrieve from, with the agent's starting position
  randomised conditional on each.

  In \cref{fig:pusht_sparsity_combined} we use $K=25$ block poses and $J=3$ pusher
  positions per pose for the Push-T sparse-observation experiments, yielding
  $75$ probed initial conditions per benchmark seed-set.

For each of these initial conditions we report mean coverage across $10$ different
maximum-step budgets, ranging over $[50, 500]$ in $50$-step intervals. The standard
Push-T benchmark fixes a single horizon ($\text{maxsteps} = 200$) \cite{florence2021implicitbehavioralcloning} and reports maximum coverage at
that cutoff, which conflates two distinct failure modes: a policy that never
makes progress and a policy that makes progress but needs more time to finish.
Sweeping the budget instead traces the full coverage-versus-time curve for each
method, so we can see not just \emph{whether} a policy recovers from an
adversarially---sparse start but \emph{how quickly}. This is the quantity that matters
in the sparse regime, where the relevant question is whether a method can eventually
recover at all once it is pushed off the demonstration manifold.
The sweep reveals a trend that the single-cutoff benchmark hides: at the standard
$\text{maxsteps} = 200$, RE4 is substantially stronger than both diffusion baselines,
reflecting faster and more reliable recovery from out-of-support scenes. As the
budget grows toward $\text{maxsteps} = 500$, however, the gap narrows---DP (both DDPM-100
and DDIM-10) progressively catches up, indicating that the diffusion policies are
largely able to reach comparable coverage given enough time, but are markedly slower
to do so from sparse initial conditions.
\medbreak
For the Robomimic and MimicGen tasks we probe the same regime with $75$
scenarios per task, selecting directly over environment \emph{seeds}. 
We sample $10{,}000$ candidate seeds, read the target object's
$(x, y, \psi)$ from each reset, and score them under a Gaussian KDE fitted to
the $200$ demonstrations' \emph{initial} object poses, lifted to
$(x, y, \cos\psi, \sin\psi)$ so that density is continuous across the $\pm\pi$
wrap. The $75$ lowest-density candidates are kept, subject to a $2.5$\,mm
minimum separation in $(x,y)$ so the selection cannot collapse into a single
tail. Every retained scenario falls below the $1$st percentile of the
demonstrations' own density on all four tasks.

Two tasks additionally require their placement region to be widened before this
is meaningful, because robosuite's stock samplers are narrow and the square
nut's is degenerate---its $x$-range spans $5$\,mm against $11.5$\,cm in $y$,
so the nut is effectively initialised on a line and a seed sweep explores one
axis only. We pad the sampling rectangle by $2$\,cm on Lift and $3$\,cm on
Square. Can and Threading
are left at their defaults, Can places into a bin, so its region is not a free
rectangle, and Threading's needle sampler is already a genuine 2-D region with a
$60^\circ$ yaw range. One consequence is worth stating plainly: the achievable
sparsity is bounded by what each environment's sampler can produce, so the probe
is much more severe on Square (median scenario log-density $-811$ against a
demonstration median of $+5.7$) than on Threading ($+4.1$ against $+6.2$).
\subsection{\textbf{Sparse Demonstration Experiment}}

As an alternative way to probe the sparse-data regime, we vary the number of
demonstrations available at training time. Specifically, we retrain both Diffusion
Policy and RE4 on Push-T (state and image) using strictly the first $N = 50$
trajectories, and additionally restrict RE4's retrieval memory to the same $N = 50$
demonstrations, so that no information leaks in through retrieval. We then evaluate
RE4, DDIM-10, and DDPM-100 over $100$ independently sampled random seeds, for both
the state and image observation settings. These results are compared directly
against checkpoints trained on the full demonstration set ($N = \text{all} = 206$)
and evaluated on the identical $100$ seeds. Across both sparse regimes---sparse
initial conditions and sparse demonstrations---RE4 degrades more gracefully than
Diffusion Policy, exhibiting a milder drop in coverage relative to its own
full-data performance.

\section{Collision Demonstration Filtering}
\label{sec:collision}
The environment is a stock \texttt{PickPlaceCan} with one immovable post added.
The post is a box with \emph{no} free joint, so it is welded into the arena. We
quantify the impact on demonstrations, as the height of the added post $z$ varies.

\begin{center}\footnotesize
\begin{tabular}{lccccccc}
\toprule
post top $z$ [m] & 0.98 & 1.02 & \textbf{1.03} & 1.04 & 1.06  \\
demos invalidated & 22.5\% & 41.5\% & \textbf{49.0\%} & 55.5\% & 72.5\%  \\
\bottomrule
\label{tab:col}
\end{tabular}
\end{center}

We take $z{=}1.03$, which invalidates $98$ of $200$ demonstrations. This means the obstacle is not an obstruction to be planned
around, instead a filter that makes roughly half the memory invalid while leaving roughly half as viable alternatives.

\subsection{Perceiving the obstacle}
The obstacle is never modelled, labelled or given to the policy. It is obtained
by \emph{subtraction} from the same RGB-D observation. RE4
knows where the target object is via pose estimation, and where its own body is
via forward kinematics, so whatever remains in the depth image is by definition
something it must not hit.
\begin{equation}
\begin{split}
\mathcal{C} = \operatorname{deproject}(\text{depth})
 &\;\ominus\; \text{target} \;\ominus\; \text{robot} \\
 &\;\ominus\; \text{support surface} \;\ominus\; \text{background}.
\end{split}
\end{equation}
Target and robot are removed \emph{in image space} from masks the perception
stack already produces, before deprojection; removing the robot with a sphere at
the end effector instead leaves the forearm and upper arm in the cloud as
phantom obstacles that move with the policy, which vetoes every candidate.
Points within $3$\,cm of the table plane are dropped.

Scene geometry post subtraction is not automatically an obstacle, every one of the $200$ demonstrations ends with the
gripper \emph{inside} a bin, so treating perceived structure as obstruction
flags all $200$ candidates (recall $100\%$, precision $49\%$---a filter that
rejects everything). What the filter actually needs is what is \textbf{novel}
relative to the workspace the demonstrations were recorded in. We therefore
subtract against a background cloud of the static workspace as it stood at
demonstration time, pooled over several obstacle-free resets so the can is not
baked in as permanent structure. This is not privileged information about the
post; it is the workspace scan any deployment already has.

The surviving points are voxel-downsampled at $6$\,mm and indexed in a KD-tree,
built once per observation so that testing a candidate demonstration is a batch
of nearest-neighbour queries rather than a brute-force sweep.

\subsection{Testing a retrieved demonstration}
\paragraph{The right body is the object, not the gripper}
A demonstration is unusable because replaying it would drive the \emph{can}
into the post, not because the gripper passes nearby. The two paths differ by
the grasp offset---which is precisely the quantity \textsc{Reframe} changes
from one demonstration to the next---so a gripper-based test rejects
demonstrations that were fine and accepts ones that bury the can. We reconstruct
the grasp from RE4's own estimate, never from simulator state,
\begin{equation}
G \;=\; (\robotstate^{\,\text{cur}})^{-1}\object^{\,\text{cur}},
\qquad
\object_k \;=\; \hat{\action}_k \cdot G ,
\end{equation}
map every pose of the reframed chunk through it, and sweep the object's
geometry along the resulting path. The path is resampled so no gap exceeds
$1$\,cm, without which a fast carry tunnels straight through a $6$\,cm post. The
test is applied in $\transfer$ only; in $\transit$ the object rests on the table
and the chunk does not carry it.

\paragraph{Operating point}
The margin and resolution were calibrated against the ground-truth MuJoCo labels
over all $200$ demonstrations, giving recall $94.9\%$ at precision $82.3\%$
(stride $2$, voxel $6$\,mm, margin $-10$\,mm). The negative margin is not a
fudge: deprojection and voxel quantisation place the perceived surface up to one
voxel \emph{outside} the true obstacle, so a zero-margin test flags grazes that
do not touch and precision falls to $63.5\%$. Recall is favoured deliberately, a
false alarm discarding one candidate out of roughly a hundred valid
alternatives while a miss executes a trajectory into the post.

\paragraph{The map is built once}
The obstacle is static; the arm is not. Rebuilding the cloud at every replan
looks more responsive but is strictly worse, because the arm occludes the post
from the single camera at exactly the moments the filter is consulted---recall
and precision fall to $56.8\%$ and $46.2\%$ when rebuilt, against $94.9\%$ and
$82.3\%$ on the unoccluded first frame. We therefore map the obstacle once from
the first frame and persist it.

\subsection{Integration with \textsc{Replan}}
The filter enters as an optional veto on retrieval and changes nothing else: when
a candidate plan is rejected it is excluded \emph{for that replan only} and
retrieval is re-run, and if every candidate is vetoed the policy falls back to
the unfiltered best so it is never left without a plan. The filter also abstains
when the gripper is already in contact at the current pose, since every candidate
then inherits the same collision and vetoing cannot discriminate between them.

\subsection{Benchmark}
Performance is benchmarked against a can diffusion policy checkpoint, across 50 independent seeds.

%% file: main.bbl
\begin{thebibliography}{10}
\providecommand{\url}[1]{#1}
\csname url@rmstyle\endcsname
\providecommand{\newblock}{\relax}
\providecommand{\bibinfo}[2]{#2}
\providecommand\BIBentrySTDinterwordspacing{\spaceskip=0pt\relax}
\providecommand\BIBentryALTinterwordstretchfactor{4}
\providecommand\BIBentryALTinterwordspacing{\spaceskip=\fontdimen2\font plus
\BIBentryALTinterwordstretchfactor\fontdimen3\font minus \fontdimen4\font\relax}
\providecommand\BIBforeignlanguage[2]{{%
\expandafter\ifx\csname l@#1\endcsname\relax
\typeout{** WARNING: IEEEtran.bst: No hyphenation pattern has been}%
\typeout{** loaded for the language `#1'. Using the pattern for}%
\typeout{** the default language instead.}%
\else
\language=\csname l@#1\endcsname
\fi
#2}}

\bibitem{mason2018toward}
M.~T. Mason, ``Toward robotic manipulation,'' \emph{Annual Review of Control, Robotics, and Autonomous Systems}, vol.~1, no.~1, pp. 1--28, 2018.

\bibitem{correia2023surveydemonstrationlearning}
A.~Correia and L.~A. Alexandre, ``A survey of demonstration learning,'' \emph{Robotics and Autonomous Systems}, vol. 182, p. 104812, 2024.

\bibitem{chi2024diffusionpolicy}
C.~Chi, Z.~Xu, S.~Feng, E.~Cousineau, Y.~Du, B.~Burchfiel, R.~Tedrake, and S.~Song, ``Diffusion policy: Visuomotor policy learning via action diffusion,'' \emph{The International Journal of Robotics Research}, 2024.

\bibitem{wang2024equivariantdiffusionpolicy}
\BIBentryALTinterwordspacing
D.~Wang, S.~Hart, D.~Surovik, T.~Kelestemur, H.~Huang, H.~Zhao, M.~Yeatman, J.~Wang, R.~Walters, and R.~Platt, ``Equivariant diffusion policy,'' in \emph{Proceedings of The 8th Conference on Robot Learning}, ser. Proceedings of Machine Learning Research, P.~Agrawal, O.~Kroemer, and W.~Burgard, Eds., vol. 270.\hskip 1em plus 0.5em minus 0.4em\relax PMLR, 06--09 Nov 2025, pp. 48--69. [Online]. Available: \url{https://proceedings.mlr.press/v270/wang25a.html}
\BIBentrySTDinterwordspacing

\bibitem{pmlr-v305-jiang25a}
\BIBentryALTinterwordspacing
S.~Jiang, X.~Fang, N.~Roy, T.~Lozano-P\'{e}rez, L.~P. Kaelbling, and S.~Ancha, ``Streaming flow policy: Simplifying diffusion/flow-matching policies by treating action trajectories as flow trajectories,'' in \emph{Proceedings of The 9th Conference on Robot Learning}, ser. Proceedings of Machine Learning Research, J.~Lim, S.~Song, and H.-W. Park, Eds., vol. 305.\hskip 1em plus 0.5em minus 0.4em\relax PMLR, 27--30 Sep 2025, pp. 238--257. [Online]. Available: \url{https://proceedings.mlr.press/v305/jiang25a.html}
\BIBentrySTDinterwordspacing

\bibitem{pmlr-v270-kim25c}
\BIBentryALTinterwordspacing
M.~J. Kim, K.~Pertsch, S.~Karamcheti, T.~Xiao, A.~Balakrishna, S.~Nair, R.~Rafailov, E.~P. Foster, P.~R. Sanketi, Q.~Vuong, T.~Kollar, B.~Burchfiel, R.~Tedrake, D.~Sadigh, S.~Levine, P.~Liang, and C.~Finn, ``Openvla: An open-source vision-language-action model,'' in \emph{Proceedings of The 8th Conference on Robot Learning}, ser. Proceedings of Machine Learning Research, P.~Agrawal, O.~Kroemer, and W.~Burgard, Eds., vol. 270.\hskip 1em plus 0.5em minus 0.4em\relax PMLR, 06--09 Nov 2025, pp. 2679--2713. [Online]. Available: \url{https://proceedings.mlr.press/v270/kim25c.html}
\BIBentrySTDinterwordspacing

\bibitem{bonnaire2026diffusion}
T.~Bonnaire, R.~Urfin, G.~Biroli, and M.~M{\'e}zard, ``Why diffusion models don’t memorize: The role of implicit dynamical regularization in training,'' \emph{Advances in Neural Information Processing Systems}, vol.~38, pp. 141\,266--141\,286, 2026.

\bibitem{2025arXiv250505787H}
C.~{He}, X.~{Liu}, G.~{Sznaier Camps}, G.~{Sartoretti}, and M.~{Schwager}, ``{Demystifying Diffusion Policies: Action Memorization and Simple Lookup Table Alternatives},'' \emph{arXiv e-prints}, p. arXiv:2505.05787, May 2025.

\bibitem{vitiello2023oneshotimitationlearningpose}
\BIBentryALTinterwordspacing
P.~Vitiello, K.~Dreczkowski, and E.~Johns, ``One-shot imitation learning: A pose estimation perspective,'' in \emph{Proceedings of The 7th Conference on Robot Learning}, ser. Proceedings of Machine Learning Research, J.~Tan, M.~Toussaint, and K.~Darvish, Eds., vol. 229.\hskip 1em plus 0.5em minus 0.4em\relax PMLR, 06--09 Nov 2023, pp. 943--970. [Online]. Available: \url{https://proceedings.mlr.press/v229/vitiello23a.html}
\BIBentrySTDinterwordspacing

\bibitem{ICLR2026_bf235a1d}
\BIBentryALTinterwordspacing
Y.~Li, N.~Darwiche, A.~Razmjoo, S.~Liu, Y.~Du, A.~Ijspeert, and S.~Calinon, ``Geometry-aware policy imitation,'' in \emph{International Conference on Learning Representations}, C.~Vondrick, B.~Hariharan, C.~Raffel, L.~Pinto, D.~Yang, and A.~Faust, Eds., vol. 2026, 2026, pp. 117\,714--117\,737. [Online]. Available: \url{https://proceedings.iclr.cc/paper_files/paper/2026/file/bf235a1d6780afd979f2f81676f43413-Paper-Conference.pdf}
\BIBentrySTDinterwordspacing

\bibitem{kuffner2016motion}
J.~Kuffner and J.~Xiao, ``Motion for manipulation tasks,'' in \emph{Springer Handbook of Robotics}.\hskip 1em plus 0.5em minus 0.4em\relax Springer, 2016, pp. 897--930.

\bibitem{hauser2010multi}
K.~Hauser and J.-C. Latombe, ``Multi-modal motion planning in non-expansive spaces,'' \emph{The International Journal of Robotics Research}, vol.~29, no.~7, pp. 897--915, 2010.

\bibitem{kingston2019decoupling}
Z.~Kingston, M.~Moll, and L.~E. Kavraki, ``Decoupling constraints from sampling-based planners,'' in \emph{Robotics Research: The 18th International Symposium ISRR}.\hskip 1em plus 0.5em minus 0.4em\relax Springer, 2019, pp. 913--928.

\bibitem{dogar2011framework}
M.~Dogar and S.~Srinivasa, ``A framework for push-grasping in clutter,'' \emph{Robotics: Science and systems VII}, vol.~1, pp. 65--72, 2011.

\bibitem{vieira2022persistent}
E.~R. Vieira, D.~Nakhimovich, K.~Gao, R.~Wang, J.~Yu, and K.~E. Bekris, ``Persistent homology for effective non-prehensile manipulation,'' in \emph{2022 International Conference on Robotics and Automation (ICRA)}.\hskip 1em plus 0.5em minus 0.4em\relax IEEE, 2022, pp. 1918--1924.

\bibitem{kingston2021using}
Z.~Kingston, C.~Chamzas, and L.~E. Kavraki, ``Using experience to improve constrained planning on foliations for multi-modal problems,'' in \emph{2021 IEEE/RSJ International Conference on Intelligent Robots and Systems (IROS)}.\hskip 1em plus 0.5em minus 0.4em\relax IEEE, 2021, pp. 6922--6927.

\bibitem{ijspeert2013dynamical}
A.~J. Ijspeert, J.~Nakanishi, H.~Hoffmann, P.~Pastor, and S.~Schaal, ``Dynamical movement primitives: learning attractor models for motor behaviors,'' \emph{Neural computation}, vol.~25, no.~2, pp. 328--373, 2013.

\bibitem{sobti2021sampling}
S.~Sobti, R.~Shome, S.~Chaudhuri, and L.~E. Kavraki, ``A sampling-based motion planning framework for complex motor actions,'' in \emph{2021 IEEE/RSJ International Conference on Intelligent Robots and Systems (IROS)}.\hskip 1em plus 0.5em minus 0.4em\relax IEEE, 2021, pp. 6928--6934.

\bibitem{Wen_2024_CVPR}
B.~Wen, W.~Yang, J.~Kautz, and S.~Birchfield, ``Foundationpose: Unified 6d pose estimation and tracking of novel objects,'' in \emph{Proceedings of the IEEE/CVF Conference on Computer Vision and Pattern Recognition (CVPR)}, June 2024, pp. 17\,868--17\,879.

\bibitem{any6d}
T.~Lee, B.~Wen, M.~Kang, G.~Kang, I.~Kweon, and K.~Yoon, ``\BIBforeignlanguage{English}{Any6d: Model-free 6d pose estimation of novel objects},'' \emph{\BIBforeignlanguage{English}{Proceedings of the IEEE Computer Society Conference on Computer Vision and Pattern Recognition}}, pp. 11\,633--11\,643, 2025.

\bibitem{shen2025efficientdiffusionmodelssurvey}
\BIBentryALTinterwordspacing
H.~Shen, J.~Zhang, B.~Xiong, R.~Hu, S.~Chen, Z.~Wan, X.~Wang, Y.~Zhang, Z.~Gong, G.~Bao, C.~Tao, Y.~Huang, Y.~Yuan, and M.~Zhang, ``Efficient diffusion models: A survey,'' \emph{Transactions on Machine Learning Research}, 2025, survey Certification. [Online]. Available: \url{https://openreview.net/forum?id=wHECkBOwyt}
\BIBentrySTDinterwordspacing

\bibitem{prasad2024consistencypolicyacceleratedvisuomotor}
A.~Prasad, K.~Lin, J.~Wu, L.~Zhou, and J.~Bohg, ``Consistency policy: Accelerated visuomotor policies via consistency distillation,'' in \emph{Robotics: Science and Systems}, 2024.

\bibitem{Pari-RSS-22}
J.~Pari, N.~Shafiullah, S.~Arunachalam, and L.~Pinto, ``{The Surprising Effectiveness of Representation Learning for Visual Imitation},'' in \emph{Proceedings of Robotics: Science and Systems}, New York City, NY, USA, June 2022.

\bibitem{2024IRAL}
N.~D. {Palo} and E.~{Johns}, ``{On the Effectiveness of Retrieval, Alignment, and Replay in Manipulation},'' \emph{IEEE Robotics and Automation Letters}, vol.~9, no.~3, pp. 2032--2039, Jan. 2024.

\bibitem{huang2018generalizedtaskparameterizedskilllearning}
Y.~Huang, J.~Silv{\'e}rio, L.~Rozo, and D.~G. Caldwell, ``Generalized task-parameterized skill learning,'' in \emph{2018 IEEE international conference on robotics and automation (ICRA)}.\hskip 1em plus 0.5em minus 0.4em\relax IEEE, 2018, pp. 5667--5674.

\bibitem{franzese2025generalizablemotionpolicieskeypoint}
G.~Franzese, R.~Prakash, C.~Della~Santina, and J.~Kober, ``Generalizable motion policies through keypoint parameterization and transportation maps,'' \emph{IEEE Transactions on Robotics}, 2025.

\bibitem{sosa2023selfsupervised3dhumanpose}
J.~Sosa and D.~Hogg, ``Self-supervised 3d human pose estimation from a single image,'' in \emph{Proceedings of the IEEE/CVF Conference on Computer Vision and Pattern Recognition}, 2023, pp. 4788--4797.

\bibitem{pan2024task}
T.~Pan, R.~Shome, and L.~E. Kavraki, ``Task and motion planning for execution in the real,'' \emph{IEEE Transactions on Robotics}, vol.~40, pp. 3356--3371, 2024.

\bibitem{shome2019towards}
R.~Shome, W.~N. Tang, C.~Song, C.~Mitash, H.~Kourtev, J.~Yu, A.~Boularias, and K.~E. Bekris, ``Towards robust product packing with a minimalistic end-effector,'' in \emph{IEEE International Conference on Robotics and Automation (ICRA)}, 2019.

\bibitem{zhou2020rot6d}
Y.~Zhou, C.~Barnes, J.~Lu, J.~Yang, and H.~Li, ``On the continuity of rotation representations in neural networks,'' in \emph{Proceedings of the IEEE/CVF conference on computer vision and pattern recognition}, 2019, pp. 5745--5753.

\bibitem{schoenberger2016sfm}
J.~L. Sch\"{o}nberger and J.-M. Frahm, ``Structure-from-motion revisited,'' in \emph{Conference on Computer Vision and Pattern Recognition (CVPR)}, 2016.

\bibitem{cnos}
V.~N. Nguyen, T.~Groueix, G.~Ponimatkin, V.~Lepetit, and T.~Hodan, ``Cnos: A strong baseline for cad-based novel object segmentation,'' in \emph{2023 IEEE/CVF International Conference on Computer Vision Workshops (ICCVW)}, 2023, pp. 2126--2132.

\bibitem{dinov2}
\BIBentryALTinterwordspacing
M.~Oquab, T.~Darcet, T.~Moutakanni, H.~Vo, M.~Szafraniec, V.~Khalidov, P.~Fernandez, D.~Haziza, F.~Massa, A.~El-Nouby, M.~Assran, N.~Ballas, W.~Galuba, R.~Howes, P.-Y. Huang, S.-W. Li, I.~Misra, M.~Rabbat, V.~Sharma, G.~Synnaeve, H.~Xu, H.~Jegou, J.~Mairal, P.~Labatut, A.~Joulin, and P.~Bojanowski, ``{DINOv2: Learning Robust Visual Features without Supervision},'' \emph{{Transactions on Machine Learning Research }}, pp. 1--31, Jan. 2024. [Online]. Available: \url{https://hal.science/hal-04376640}
\BIBentrySTDinterwordspacing

\bibitem{sam}
A.~{Kirillov}, E.~{Mintun}, N.~{Ravi}, H.~{Mao}, C.~{Rolland}, L.~{Gustafson}, T.~{Xiao}, S.~{Whitehead}, A.~C. {Berg}, W.-Y. {Lo}, P.~{Dollar}, and R.~{Girshick}, ``{Segment Anything},'' in \emph{2023 IEEE/CVF International Conference on Computer Vision (ICCV)}.\hskip 1em plus 0.5em minus 0.4em\relax IEEE, Oct. 2023, p. 371.

\bibitem{lavalle2001randomized}
S.~M. LaValle and J.~J. Kuffner~Jr, ``Randomized kinodynamic planning,'' \emph{The international journal of robotics research}, vol.~20, no.~5, pp. 378--400, 2001.

\bibitem{florence2021implicitbehavioralcloning}
P.~Florence, C.~Lynch, A.~Zeng, O.~A. Ramirez, A.~Wahid, L.~Downs, A.~Wong, J.~Lee, I.~Mordatch, and J.~Tompson, ``Implicit behavioral cloning,'' in \emph{Conference on robot learning}.\hskip 1em plus 0.5em minus 0.4em\relax PMLR, 2022, pp. 158--168.

\bibitem{robomimic2021}
A.~Mandlekar, D.~Xu, J.~Wong, S.~Nasiriany, C.~Wang, R.~Kulkarni, L.~Fei-Fei, S.~Savarese, Y.~Zhu, and R.~Mart\'{i}n-Mart\'{i}n, ``What matters in learning from offline human demonstrations for robot manipulation,'' in \emph{Conference on Robot Learning (CoRL)}, 2021.

\bibitem{pmlr-v229-mandlekar23a}
\BIBentryALTinterwordspacing
A.~Mandlekar, S.~Nasiriany, B.~Wen, I.~Akinola, Y.~Narang, L.~Fan, Y.~Zhu, and D.~Fox, ``Mimicgen: A data generation system for scalable robot learning using human demonstrations,'' in \emph{Proceedings of The 7th Conference on Robot Learning}, ser. Proceedings of Machine Learning Research, J.~Tan, M.~Toussaint, and K.~Darvish, Eds., vol. 229.\hskip 1em plus 0.5em minus 0.4em\relax PMLR, 06--09 Nov 2023, pp. 1820--1864. [Online]. Available: \url{https://proceedings.mlr.press/v229/mandlekar23a.html}
\BIBentrySTDinterwordspacing

\bibitem{wolberg_robust}
G.~Wolberg and S.~Zokai, ``Robust image registration using log-polar transform,'' in \emph{Proceedings 2000 International Conference on Image Processing (Cat. No.00CH37101)}, vol.~1, 2000, pp. 493--496 vol.1.

\bibitem{he2015deepresiduallearningimage}
K.~He, X.~Zhang, S.~Ren, and J.~Sun, ``Identity mappings in deep residual networks,'' in \emph{European conference on computer vision}.\hskip 1em plus 0.5em minus 0.4em\relax Springer, 2016, pp. 630--645.

\bibitem{shi2023zero123plus}
R.~{Shi}, H.~{Chen}, Z.~{Zhang}, M.~{Liu}, C.~{Xu}, X.~{Wei}, L.~{Chen}, C.~{Zeng}, and H.~{Su}, ``{Zero123++: a Single Image to Consistent Multi-view Diffusion Base Model},'' \emph{arXiv e-prints}, p. arXiv:2310.15110, Oct. 2023.

\bibitem{xu2024instantmesh}
J.~Xu, W.~Cheng, Y.~Gao, X.~Wang, S.~Gao, and Y.~Shan, ``Instantmesh: Efficient 3d mesh generation from a single image with sparse-view large reconstruction models,'' \emph{arXiv preprint arXiv:2404.07191}, 2024.

\bibitem{byol}
\BIBentryALTinterwordspacing
J.-B. Grill, F.~Strub, F.~Altch\'{e}, C.~Tallec, P.~Richemond, E.~Buchatskaya, C.~Doersch, B.~Avila~Pires, Z.~Guo, M.~Gheshlaghi~Azar, B.~Piot, k.~kavukcuoglu, R.~Munos, and M.~Valko, ``Bootstrap your own latent - a new approach to self-supervised learning,'' in \emph{Advances in Neural Information Processing Systems}, H.~Larochelle, M.~Ranzato, R.~Hadsell, M.~Balcan, and H.~Lin, Eds., vol.~33.\hskip 1em plus 0.5em minus 0.4em\relax Curran Associates, Inc., 2020, pp. 21\,271--21\,284. [Online]. Available: \url{https://proceedings.neurips.cc/paper_files/paper/2020/file/f3ada80d5c4ee70142b17b8192b2958e-Paper.pdf}
\BIBentrySTDinterwordspacing

\bibitem{ross2011reductionimitationlearningstructured}
S.~Ross, G.~Gordon, and D.~Bagnell, ``A reduction of imitation learning and structured prediction to no-regret online learning,'' in \emph{Proceedings of the fourteenth international conference on artificial intelligence and statistics}.\hskip 1em plus 0.5em minus 0.4em\relax JMLR Workshop and Conference Proceedings, 2011, pp. 627--635.

\bibitem{sridhar2024memoryconsistentneuralnetworksimitation}
K.~Sridhar, S.~Dutta, D.~Jayaraman, J.~Weimer, and I.~Lee, ``Memory-consistent neural networks for imitation learning,'' in \emph{International Conference on Learning Representations}, vol. 2024, 2024, pp. 45\,160--45\,185.

\end{thebibliography}
